%% file: main.tex
\documentclass{article}

\usepackage[preprint]{neurips_2023}

\usepackage[utf8]{inputenc} %
\usepackage[T1]{fontenc}    %
\usepackage{hyperref}       %
\usepackage{url}            %
\usepackage{booktabs}       %
\usepackage{amsfonts, amsmath, amssymb, mathrsfs, mathtools}       %
\usepackage{nicefrac}       %
\usepackage{microtype}      %
\usepackage{xcolor}         %
\usepackage{csquotes}       %
\usepackage{graphicx}
\usepackage{subcaption}
\usepackage[most]{tcolorbox}
\usepackage{tikzscale}
\usepackage{soul}
\usepackage{listings}
\usepackage{tikz}
\usepackage{svg}
\usepackage{epsfig}
\usepackage{xspace}
\usepackage{algorithm}
\usepackage{algorithmic}
\usepackage[font=footnotesize]{caption}
\usepackage[export]{adjustbox}

\tcbuselibrary{breakable, skins}
\usepackage{spverbatim}

\graphicspath{{shrunk_figures/}}

\hypersetup{
    colorlinks=true,
    linkcolor=blue,
    citecolor=blue,
    urlcolor=blue,
}
\DeclareMathOperator{\dataset}{\mathcal{D}}

\newcommand{\centermatrix}{H}
\newcommand{\adjustedE}{\mathcal{E}}

\newcommand{\LB}{Lucius Bushnaq\xspace}
\newcommand{\JM}{Jake Mendel\xspace}
\newcommand{\SH}{Stefan Heimersheim\xspace}
\newcommand{\NGD}{Nix Goldowsky-Dill\xspace}
\newcommand{\DB}{Dan Braun\xspace}
\newcommand{\MH}{Marius Hobbhahn\xspace}

\newcommand{\KH}{Kaarel Hänni\xspace}
\newcommand{\AG}{Avery Griffin\xspace}
\newcommand{\JS}{Jörn Stöhler\xspace}

\title{The Local Interaction Basis: Identifying Computationally-Relevant and Sparsely Interacting Features in Neural Networks}

\author{%
  Lucius Bushnaq\thanks{Correspondence to Lucius Bushnaq <lucius@apolloresearch.ai>}\quad
  Stefan Heimersheim\quad
  Nicholas Goldowsky-Dill\quad
  Dan Braun\thanks{Lead Engineer}\quad
  Jake Mendel\\
  \AND %
  Kaarel Hänni\thanks{Cadenza Labs}\quad
  Avery Griffin\thanks{Independent}\quad
  Jörn Stöhler \footnotemark[4]\quad
  Magdalena Wache \footnotemark[4]\quad
  \AND %
  Marius Hobbhahn\thanks{See Section \ref{sec:contribution} for contributions} 
  \AND \textmd{Apollo Research}
}

\begin{document}

\maketitle

\setcounter{footnote}{0}
\begin{abstract}
Mechanistic interpretability aims to understand the behavior of neural networks by reverse-engineering their internal computations. However, current methods struggle to find clear interpretations of neural network activations because a decomposition of activations into computational features is missing. Individual neurons or model components do not cleanly correspond to distinct features or functions. We present a novel interpretability method that aims to overcome this limitation by transforming the activations of the network into a new basis - the Local Interaction Basis (LIB).
LIB aims to identify computational features by removing irrelevant activations and interactions. Our method
drops irrelevant activation directions and aligns the basis with the singular vectors of the Jacobian matrix between adjacent layers. It also scales features based on their importance for downstream computation, producing an interaction graph that shows all computationally-relevant features and interactions in a model.
We evaluate the effectiveness of LIB on modular addition and CIFAR-10 models, finding that it identifies more computationally-relevant features that interact more sparsely, compared to principal component analysis. However, LIB does not yield substantial improvements in interpretability or interaction sparsity when applied to language models. We conclude that LIB is a promising theory-driven approach for analyzing neural networks, but in its current form is not applicable to large language models.

\end{abstract}

\newcounter{boxnumber}
\newcounter{prompt}

\input{sections/introduction}

\input{sections/methodology}
\input{sections/results}

\input{sections/conclusions}

\bibliography{references}
\bibliographystyle{plainnat}

\appendix
\clearpage
\input{sections/appendix}

\end{document}

%% file: sections/introduction.tex
\section{Introduction}
\label{section/intro}
Mechanistic Interpretability aims to understand the internals of neural networks and to reverse
engineer computation inside neural networks \citep{olah2017feature,elhage2021mathematical}.
Previous attempts have analyzed toy models \citep{chughtai2023toy, nanda2023progress} or circuits that compute specific subtasks
performed by large language models \citep{olah2020zoom, meng2023locating, geiger2021causal, wang2022interpretability, conmy2024towards}.
These analyses largely rely on interpretations of model components \citep{wang2022interpretability},
individual neurons \citep{gurnee2023finding}, or principal components \citep{millidge2022singular}.
Other approaches to disentangling the features learned in the latent spaces of the network include
\cite{schmidhuber1992learning, desjardins2012disentangling, kim2018disentangling, chen2016infogan, peebles2020hessian, schneider2021explaining}; see \citet{bengio2014representation} for a review.

However, it has become clear that the standard basis (aligned with activation basis directions, sometimes referred to as ``neuron basis'')
is not the right unit for interpretability due to polysemanticity: neurons \citep{olah2017feature,nguyen2016,goh2021multimodal,geva2021}
and model components \citep[e.g. attention heads,][]{Janiak_Mathwin_Heimersheim_polysemantic} do not correspond to individual features.
Yet there is evidence that features are linearly represented in the activation space of neural networks
\citep{nanda2023progress,gurnee2023finding,nanda2023emergent}.
Therefore, there are two possibilities for how features are represented in neural networks:
(a) the model has more features than dimensions, but these features are sparsely activating represented in superposition
\citep{elhage2022superposition,Sharkey_Braun_Millidge_2022,Vaintrob_Mendel_Kaarel_2024},
or (b) features are represented in a non-overcomplete basis, but not aligned with neurons.\footnote{Intermediate options are also a possibility: For example, the model's activation space could be split into different subspaces, some of which contain more sparsely activating features than the dimension of the subspace, while others contain non-sparse features and do not have more features than dimensions.}
While option (a) has some merits \citep[interpretability results, see][]{cunningham2023sparse,bricken2023monosemanticity,marks2024sparse},
its drawback is a higher complexity. We therefore focus on testing option (b) and present a
method that assumes a non-overcomplete basis of features.

\begin{figure}[t]
    \centering
    \includegraphics[width=\linewidth]{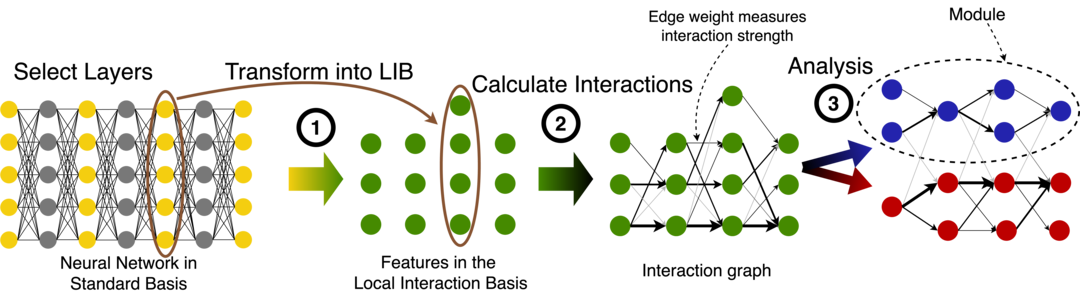}
    \caption{The Local Interaction Basis (LIB) is a basis for neural network activations where interactions between features should be sparser and more modular. (1) We start with a selection of layers from the neural network. (2) We transform the activations in these layers into the LIB, which represents computationally-relevant features, removes features that don't affect the output, and minimizes interactions between features in adjacent layers. (3) We then quantify the interactions between features using integrated gradients, creating an interaction graph that represents the extent to which preceding nodes affect subsequent nodes. (4) We use the resulting interaction graph to analyze and interpret features in the neural network, and to identify modules that correspond to distinct circuits in the model's computation.}
    \label{fig:rib_overview}
\end{figure}

Our work introduces two novel contributions to the field of mechanistic interpretability. First, we develop the Local Interaction Basis (LIB), a method for finding a more interpretable basis for neural network activations. LIB builds upon the theoretical framework proposed by \citet{apolloTheory}, which aims to find a parameterization-invariant representation of neural networks. The key idea
is that the standard basis representation of a neural network's parameters contains superfluous structure that hinders interpretability. Our method removes degenerate directions in layer activations and in gradients between adjacent layers.
This yields simpler but equivalent representation of the network, containing only computationally-relevant features---features that are relevant for
downstream computation. The LI basis is aligned with singular vectors of the Jacobian between layers such that LIB features are sparsely interacting, and ordered by their effect on the next layer.

Second, we propose the use of integrated-gradient interaction graphs to analyze the relationships between features in the LIB-transformed network. Integrated gradients \citep[IGs,][]{friedman2004} have been previously used to attribute neural network outputs to inputs \citep{sundararajan2017axiomatic}, and recently been applied
to sparse autoencoder features \citep{marks2024sparse}. We employ IGs to represent
the full network as an interaction graph to reveal hidden structure in neural networks.
IGs are particularly well-suited for this purpose due to their desirable properties, including implementation invariance, completeness, sensitivity, linearity \citep{sundararajan2017axiomatic}, and robustness to basis transformations \citep{apolloIG}.

We apply our method to two toy models and two language models. We transform activations into the LI basis and summarize the network as an interaction graph. We successfully isolate computationally-relevant features in the modular addition and CIFAR-10 toy models and are able to interpret model features based on the interaction graphs. We find that the method does not work well on language models (Tinystories-1M and GPT2-small): while the LI basis is more sparsely-interacting than the PCA baseline in some cases, the interactions remain relatively dense and the features are no more interpretable.

In this paper, we describe the LIB method, our IG-based interaction graph, and our graph analysis
methods (Section~\ref{section/methodology}). We apply the new tools to a modular addition transformer,
an MLP trained on CIFAR 10, and two language models (GPT2-small and Tinystories-1M) and present the
results in Section~\ref{sec:experiments}. We conclude in Section~\ref{sec:conclusion}.

%% file: sections/methodology.tex
\section{Methodology}
\label{section/methodology}
\begin{figure}[t]
    \centering
    \includegraphics[width=\linewidth]{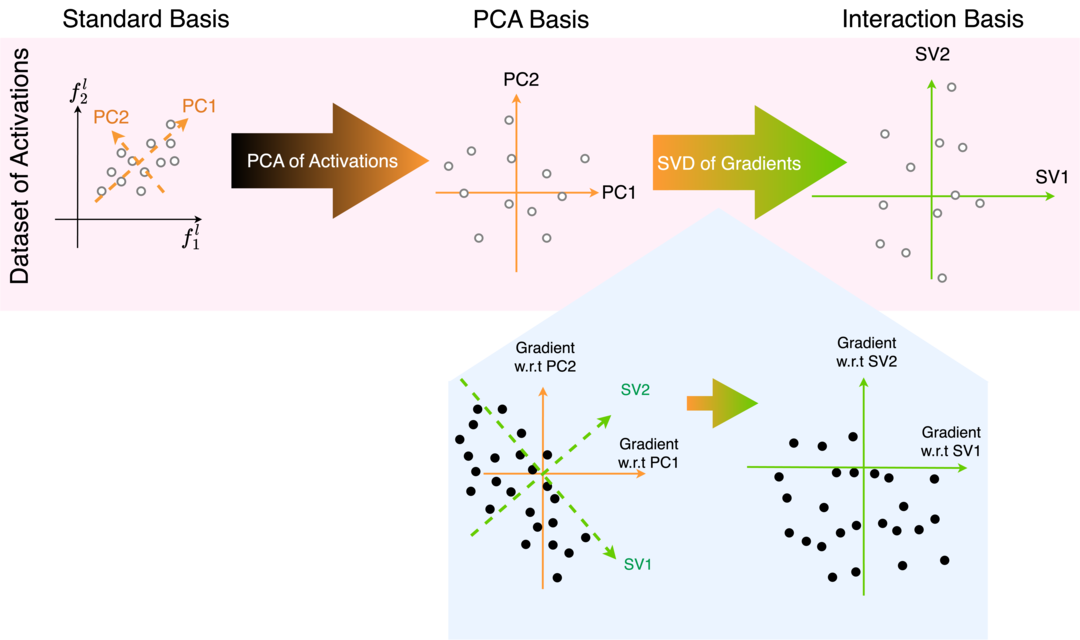}
    \caption{
    Visualization of the LIB transformation. This figure shows an illustration of
    activations (top) and gradients (bottom) as they get transformed into the LIB.
    The first step is a PCA of the activations in every layer in order to drop
    activation directions with near-zero variance and to whiten the activations.
    The second step is based on a dataset of gradients, that is, the set of gradients of every
    feature in the next layer with respect to every direction in the current layer on every data
    point (this is a larger dataset than the activations).
    We perform an SVD (singular value decomposition) on the Jacobians to find the right singular vectors and singular values.
    This allows us to drop directions that are not important for the next layer, and to align the
    activations singular vectors to sparsify the interactions between features in adjacent layers.
    }
    \label{fig:rib_rotation}
\end{figure}

Our interpretability method represents a network in a new basis which better captures its computational structure. 
We expect that these new basis directions correspond
to meaningful features of the model, enabling us to interpret individual features and feature interactions across the network.
Our method involves three key steps: first, transforming activations into a local interaction basis (LIB, Section \ref{subsec:LIB}); second, computing integrated-gradient attributions (Section \ref{subsec:IG}); and third, creating and analyzing an interaction graph to identify modules (Section \ref{subsec:modules}).
Figure \ref{fig:rib_overview} provides an overview of this process, below we provide details on each step.
\begin{enumerate}
    \item[]{\textbf{Select layers to analyze:}
    The first step is to choose a subset of layers of the neural network to include in the LIB interaction graph. To see the connection structure of the network, graph layers should usually be chosen close to each other, e.g. after every attention and MLP layer in a transformer. We index layers of the LIB graph with $l = 1,\dots,l_\text{final}$.
    }
    \item{\textbf{Transform into local interaction basis:}
    We apply a linear transformation to bring the activations into the LI basis (illustrated in Figure \ref{fig:rib_rotation}):
    \begin{enumerate}
        \item Transformation into the PCA (principal component analysis) basis: In each layer, we calculate the principal
        components of the activation vectors collected over the dataset and transform into a basis
        aligned with these principal components.
        \item Transformation to the LI basis: We iterate through the chosen network layers, in order
        from the closest to the outputs to the closest to the inputs. In every layer, we compute how
        much every direction in a layer ``connects'' to every direction in the following layer by
        calculating a Jacobian matrix: the gradients of LIB features in the following layer with
        respect to directions in the current layer. Then we transform into a basis aligned with the
       right-handed singular vectors of the Jacobian.
    \end{enumerate}
    }
    \item \textbf{Calculate interaction edges:} We build an \emph{interaction graph}, a graph of all
    LIB features in the network with edges representing the strength of interaction between features
    in adjacent layers. The interaction strength between two features is computed by calculating
    integrated gradients-based attributions \citep{sundararajan2017axiomatic} on every data point
    and then taking the quadratic mean over the dataset.
    \item{\textbf{Analyze the graph:}
    We test the sparsity of interactions (by ablating interactions), cluster the graph into modules, and interpret the nodes.}
    \end{enumerate}
    
\subsection{Step 1: The Local Interaction Basis (LIB)}
\label{subsec:LIB}

In this section, we describe the LIB transformation.
LIB is based on a predecessor, the (global) interaction basis described in \citet{apolloTheory} and
Appendix \ref{appendix/gradient_flow}; here we
describe the local interaction basis and provide
a concrete implementation.
The two types of architectures we consider here are MLPs and transformers (see Appendices
\ref{appendix/models_modular} through \ref{appendix/models_language} for model details).
In both cases we reformulate the architectures such that they can be written
as a sequential composition of layers (concatenating the residual stream and component activations of transformers), and such
that every layer returns zero if its input is zero (we fold-in the biases).
We describe these adjustments in Sections \ref{appendix/sequential_transformer}
and \ref{appendix/bias_fold}, respectively.

The codebase implementing LIB and our integrated gradient computation
is available at \url{https://github.com/ApolloResearch/rib}, and we provide
the pseudocode for the basic algorithms in Appendix \ref{appendix/pseudocode}.

The transformation $\hat{\mathbf{f}}^{l} = C^{l} \mathbf{f}^l$ to change into the LIB is best understood as consisting of
two sequential linear transformations: a transformation into the PCA basis of the activations, followed by
a transformation into the basis of right singular vectors of the Jacobian matrix to
the next layer.

The first transformation into the PCA basis consists of four steps. We
(i) center the activations over the dataset and calculate principal components and values,
and (ii) rotate the activations into a basis aligned with its principal components.
Then we (iii) drop directions with near-zero principal values, and (iv) rescale the remaining
directions with a diagonal matrix such that their covariance matrix is the
identity. We give a precise mathematical description in appendix \ref{appendix/first_transformation}.

The purpose of this first transformation is mainly to drop irrelevant directions
(those with near-zero variance), and to whiten the activations as a preparation
for the second transformation.

For the second transformation we compute the Jacobians $J^{l}_{ij}(x)$, the gradient of the $i$-th
feature in layer $l+1$ with respect to the $j$-th feature in layer $l$ for every data point $x$.
We then compute the singular value decomposition (SVD) of the $n_{\text{data}} d^{l+1} \times d^{l}$
dimensional all-data Jacobian matrix (a flattening of the 3-dimensional Jacobian tensor along the $x$ and $i$ indices)
and save the right singular vectors and singular values.
One might think of this as taking a PCA of the set of all the Jacobian rows (for all $i$
and $x$) except without centering. We (i) rotate the activations into a basis aligned with
these right singular vectors, (ii) drop directions with near-zero eigenvalues,
and (iii) rescale the activations with corresponding singular values. We give
a precise mathematical description in Appendix \ref{appendix/second_transformation}.

The second rotation achieves multiple goals. Firstly, we can drop directions in activation space that
explain variance in the current layer (not dropped by first rotation) but have no influence on
future layers (zero singular values). Secondly, the rotation into the SVD basis should
make the interactions between features in adjacent layers as sparse as possible.\footnote{The
SVD basis for a \textit{single} datapoint always gives the sparsest interactions. In our case
we choose the SVD of the reshaped matrix, inspired by the procedure for higher-order SVD, as
a guess of the best basis for the full dataset. We have not proven that this basis leads to
sparsest interactions in all cases.}
Finally, the multiplication by the singular values scales the features proportional to
how important they are for the following layers.

This is a recursive process, starting from the final layer and working backwards. In that final
layer we start the recursion by only applying the PCA step. Specifically we only perform mean-centering
and alignment with PCA directions but not the rescaling.
In addition to the recursion-based \textit{local} interaction basis we also considered a global
interaction basis where we apply the second transformation just with respect to the final layer,
rather than the next layer. We describe this alternative approach, closer to the version proposed in \citet{apolloTheory}, in appendix
\ref{appendix/gradient_flow}. We found that the results were similar, and decided to focus
on LIB due to its lower computational cost.

Our transformer implementation needs to account for the token indices, in addition to the feature indices.
This increases the computational cost of computing the Jacobians (which now depend on four rather than two
indices). For this reason we introduce a approximation technique based on stochastic sources \citep[][]{Dong_1994}
which we find to provide accurate results in practice. We provide the equations
for the transformer implementation in Appendix \ref{appendix/transformer_implementation},
and describe the stochastic sources technique in Appendix \ref{appendix/stoch_sources}.

\subsection{Step 2: Quantifying interactions with integrated gradients}
\label{subsec:IG}
In the previous subsection we derived a basis for the activations to identify the features in every layer.
In this subsection we quantify the interaction strength between features in different layers.

We use integrated gradients \citep{friedman2004,sundararajan2017axiomatic} to attribute the influence of one
feature onto another. The reason we choose this method it that it uniquely satisfies a set of properties
we want from an attribution method: implementation invariance, completeness, sensitivity, linearity,
and consistency under coordinate transformations
\citep{apolloIG}.

Integrated gradients yield an attribution $A_{i,j}^{l+1,l}(x)$ quantifying the influence of
feature $f^l_j$ in layer $l$ on feature $f^{l+1}_i$ in layer $l+1$, at a given data point $x$:
\begin{equation}\label{eq:datapoint_attr}
A^{l+1,l}_{ij}(x)\coloneq 
    f^{l}_j(x)\int^1_{0} \text{d}\alpha 
        \left[\frac{
            \partial
        }{
            \partial z^l_j
        } (F^{l+1,l}_i(\mathbf{z}^l))
        \right]_{\mathbf{z}^l=\alpha \mathbf{f}^{l}(x)}
\end{equation}
where $F$ is the function that maps $\mathbf{f}^l$ to $\mathbf{f}^{l+1}$,
$F(\mathbf{f}^l(x)) = \mathbf{f}^{l+1}(x)$.

To obtain averaged attributions $ E^{l+1,l}_{i,j}$for the entire dataset, we take the RMS of the attributions for
individual data points.
We choose the RMS as the simplest way to average over the dataset\footnote{The simplest way beyond simply summing attributions; a simple sum would lead to undesired cancellations between positive and negative attributions.} but other ways to average attributions may be considered in future work.

To speed up the computation of the attributions, we use the fact that the integral in equation
\eqref{eq:datapoint_attr} is linear in $\mathbf{z}^l$ for elementwise activation functions
(such as ReLU). Thus we can skip the integration in such cases, and just evaluate the integrand
at $\alpha=1$. For approximately linear activation functions (such as GELU) we find that
this is still a good approximation and thus use this approximation throughout.

For transformer models, the activations vary with token index and data point, and activations
of later tokens can depend on previous tokens. We generalize equation \eqref{eq:datapoint_attr}
by considering  the influence of $f^{l}_{j,t}(x)$ on $f^{l+1}_{i,s}(x)$
(with token indices $t$ and $s$). Instead of taking the RMS over the dataset, we take the RMS
summing over both, dataset index $x$ and token index $s$. The dataset- and token-averaged edge attribution formula for transformers is thus
\begin{equation}\label{eq:edges_transformers}
\begin{aligned}
E^{l+1,l}_{i,j} \coloneq \left(\frac{1}{\vert \dataset \vert}\frac{1}{T}\sum_{x\in \dataset}\sum^T_{s=1} {\left(\sum^T_{t=1}f^{l}_{j,t}(x)\int^1_{0}\text{d}\alpha \left[\frac{\partial}{\partial z^l_{j,t}} \left(F^{l+1,l}_{i,s}(\mathbf{z}^l)\right)\right]_{\mathbf{z}^l=\alpha \mathbf{f}^{l}(x)}\right)}^2\right)^{1/2}.
\end{aligned}
\end{equation}

As discussed in Section \ref{subsec:LIB}, calculating Jacobians for all indices $i, j, s, t$
is computationally expensive and we apply stochastic sources to reduce the computational cost
(see Appendix \ref{appendix/stoch_sources}).

\subsection{Step 3: Interaction graph analysis}
\label{subsec:interaction_graphs}
Based on the features and attributions between features, we create a layered graph. The nodes in the graph are the
features in the network $\hat f^{l}_j$ for $j = 1,\dots,d^l$ in layer $l$, and the edges between
a pair of features in adjacent layers are given by the averaged attributions $\hat E^{l+1,l}_{i,j}$.\footnote{We are using $\hat E$ to denote the attribution between LIB features $\hat{\mathbf{f}}$, rather than standard basis features.} We show
examples of these graphs in Section \ref{sec:experiments}.

\label{method:edge_ablation}
We test the \textit{sparsity} of the interactions by running edge ablations, implemented as
follows: To measure the effect of ablating the edge $\hat E^{l+1,l}_{i_0,j_0}$, we compute a forward pass
$l \rightarrow {l+1}$ setting node $\hat{\mathbf{f}}^{l}_{j_0}$ to 0, and saving only the result for
$\hat{\mathbf{f}}^{{l+1}}_{i_0}$. Then, we compute another forward pass $l \rightarrow {l+1}$ without
ablating $\hat{\mathbf{f}}^{l}_{j_0}$, saving the results for $\hat{\mathbf{f}}^{{l+1}}_{i},\, i \neq i_0$.
Finally, we combine these results to obtain the activations for layer $l+1$ with the edge ablated,
and pass them through the rest of the network. We can also ablate multiple edges at once by ablating a different
subset of input nodes when computing the value of each output node. In that case we perform one forward pass for
each node in layer $l+1$ that requires a different
input mask (up to $d^{l+1}$ forward passes).
As with most activation patching experiments \citep{vig2020causal,geiger2021causal,meng2023locating,goldowskydill2023localizing}, there is a risk that this method takes the network
out of distribution in a somewhat unprincipled manner \citep{chan2022causal}, but the technique is commonly used and
seems to work in practice.

To test the sparsity of a layer in a given basis we sort all edges by their size $\hat E^{l+1,l}_{i,j}$ and
then remove as many edges as possible, starting with the smallest ones, while maintaining a given
accuracy or loss. We implement this as a bisect search, and typically require on the order of 10-20
iterations to find the number of edges required to maintain the given accuracy/loss. This method
relies on the edge size being a good proxy for the importance of an interaction, otherwise an important edge may be ablated while less important edges are kept. 

\label{subsec:modules}
Finally, we aim to find modules in the interaction graph as a way to identify circuits in the computation
of the network. We are interested in modularity in the sense of graph sections that have comparatively low \textit{node-connectivity} with the rest of the network. That is, the minimum number of nodes
that need to be removed from the graph to disconnect two modules. However, we cannot cheaply compute
the node-connectivity of the graph, so we use a modularity score which measures how many edges
connect nodes in the same module compared to how many edges connect nodes in different modules.
This is an approximation, and we see our modularity analysis as a first proof of concept rather than
the definitive method. For more details see \citet{apolloTheory}.

In practice, we used the Leiden algorithm \citep{Traag_2019} for its speed and scalability. To
emulate the effect of the node-connectivity, we set the edges passed to the Leiden algorithm to
\begin{equation}
    \adjustedE^{l+1,l}_{i,j} = \log\left(\hat E^{l+1,l}_{i,j}/\epsilon\right)
\end{equation}
where $\epsilon$ is the smallest edge required
to maintain a high level of accuracy (obtained via edge ablation experiments). The reasoning for
the logarithmic scaling follows \citet{apolloTheory}.
The Leiden algorithm optimizes the community assignments $c^l_i$ to maximize the modularity score
\begin{equation}
Q=\frac{1}{2m}\sum_{a,i,j} \left(\adjustedE^{l+1,l}_{i,j}-\frac{k^{{l+1}}_i k^{l}_j}{2m} \right)\delta({c^{{l+1}}_i,c^{l}_j})\,.
\end{equation}
where $k^{l+1}_i=\sum_j \adjustedE^{l+1,l}_{i,j}$ and $m=\sum_{i,j} \adjustedE^{l+1,l}_{i,j}$.

%% file: sections/results.tex
\section{Results}
\label{sec:experiments}
In this section, we show experimental results for a transformer trained on a modular addition task, a CIFAR-10 MLP, and two
language models. We compare the LI basis to the PCA basis as a baseline.\footnote{PCA suggests itself as a minimal baseline since LIB aims to be an improved version of PCA. However, SAEs are the interpretability method that currently yields the most human-interpretable results. Subjectively, we found that features found by SAEs are more human-interpretable than those found by LIB.}
We find the following:
\begin{enumerate}
    \item LIB features are about as interpretable as PCA features (Sections
    \ref{subsec:modular_addition} and \ref{subsec:llms}), and more computationally-relevant (in the cases discussed in Sections \ref{subsec:modular_addition} and \ref{subsec:cifar10}).
    \item Integrated gradient interaction graphs and modularity analyses can be useful for circuit
    discovery (Section \ref{subsec:modular_addition} and \ref{subsec:cifar10}).
    \item LIB features are more sparsely interacting than PCA features in CIFAR-10 (Section \ref{subsec:cifar10})
    and most layers of the language models we tested (Section \ref{subsec:llms}).
\end{enumerate}

\subsection{Modular Addition Transformer}
\label{subsec:modular_addition}
Modular addition is a well-studied task on which transformer models develop simple, human-interpretable algorithms \citep{nanda2023progress,chughtai2023groupoperations,zhong2023pizza}.
The task is to predict the result of the mathematical operation $z = x+y \mod{p}$
(for a fixed $p=113$ in our case). We train a transformer model on this task,
see Appendix \ref{appendix/models_modular} for details.
Transformers tend to implement the \enquote{clock} algorithm
\citep{nanda2023progress} which calculates $\cos(\omega(x+y+\phi))$ for various values of $\omega$
and $\phi$ and combines those to calculate $z$. A simple version of this algorithm would be to
make the logit for output token $a$ proportional to
$\cos(2\pi(x+y-a)/p)$, for all outputs. Then the highest logit would always be the one where
$a = z = x+y\mod p$. In practice, models tend to use multiple frequencies $< p$ to improve the confidence of the prediction.

This algorithm is particularly easy to visualize because we can use a Fourier transform to represent
any neural network activation as a sum of sinusoidal terms of different frequencies (see Appendix \ref{appendix/modular_addition_math} for details). In short, we decompose any
feature into terms such as 70\% $\cos(7x+7y)$ + 30\% $\cos(31x+31y)$ which means that 70\% (and
30\%) of the variance in that feature's value can be explained by an oscillating function with frequency 7 (and
31) in the $x+y$ direction. We use this notation to label LIB features and PCA features in the
interaction graphs below.

We test whether LIB and PCA identify meaningful computational features of the algorithm, and how they compare. We do this by checking
\begin{itemize}
    \item \textbf{Functional relevance:} All features found should be computationally-relevant to the output of the network. We check
    this based on their interactions with future layers.
    \item \textbf{Monosemanticity:} Features should mostly represent single Fourier frequencies.\footnote{We would expect
    good features before the attention layer to be
    largely mono-frequency because taking the product of two frequencies results in cross-terms
    that are not helpful for the modular addition algorithm. In later layers, in particular in the
    unembedding layer, features don't necessarily have to be mono-frequency anymore.}
    \item \textbf{Sparsity:} Features should interact sparsely, i.e. there should be few edges of relevant size in our interaction graph.
    We operationalize this by checking how many edges we can ablate while retaining $>99.9\%$ test accuracy.
    \item \textbf{Modularity:} Is there modular structure in the interaction graph? For instance, is there a set
    of features that track a single set of frequencies and which interact much more strongly with each other than with other sets of features?\footnote{Note that we expect up to two
    pure features for every frequency. This is because there can be two independent terms
    that differ in their phase $\phi$ but we omit this phase in the labeling.}
\end{itemize}

We visualize results as an interaction graph as introduced in Section \ref{subsec:interaction_graphs}.
Nodes in the graph are LIB features, split by layer, sorted by importance, and colored by module.
In the modular addition interaction graphs, we set the edge line widths to be proportional
to the squared interactions $\hat E^2_{ij}$ to match the explained variance description. We also
normalize the edges sizes per-layer to improve readability for the graph in the PCA basis.

\begin{figure}[ht]
    \centering
    \adjincludegraphics[width=\textwidth,trim={0 0.23\height{} 0 0.03\height},clip]{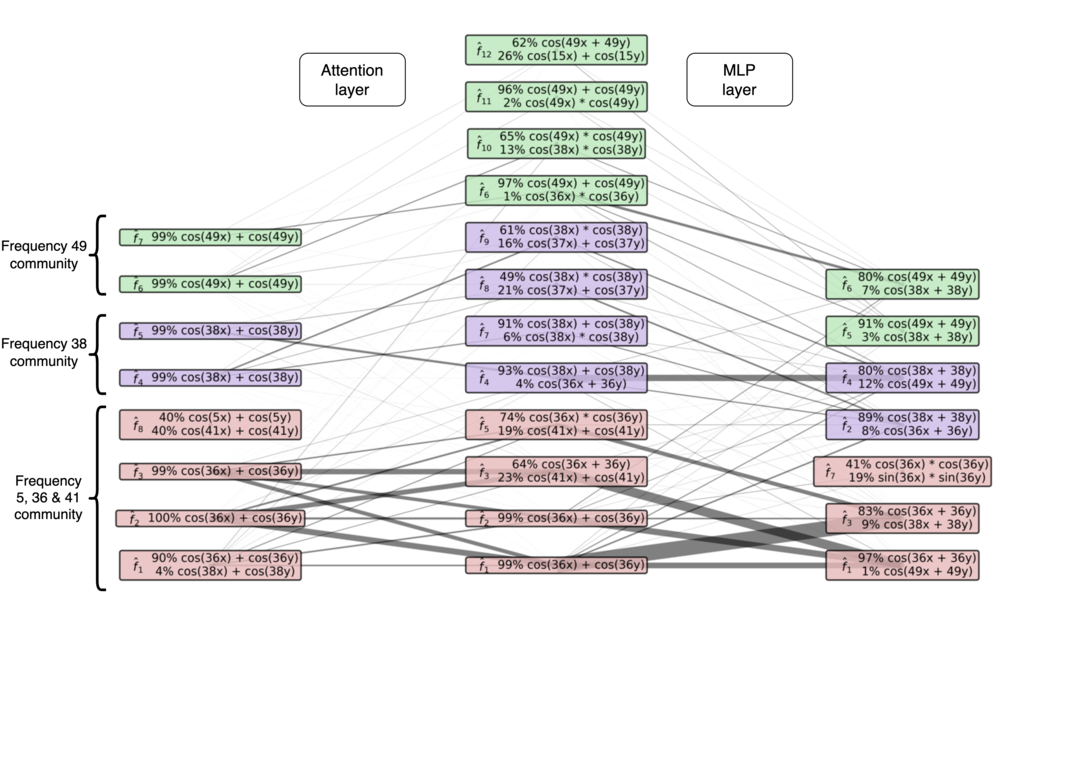}
    \caption{LIB interaction graph of a modular addition transformer. 
    The three layers correspond to
    activations after the embedding, directly after the attention, and just
    before the unembedding. 
    The individual nodes represent LIB features, and the thickness of the edges shows the interaction strength between features. The nodes are colored by module membership (Leiden algorithm),
    and labeled by their function index ($\hat f_0, \hat f_1,\dots$, in order of decreasing functional importance) and their Fourier interpretation.}
\label{fig:modular_arithmetic_rib_graph_cherry_picked}
\end{figure}

We show the LIB interaction graph of a modular addition model in Figure
\ref{fig:modular_arithmetic_rib_graph_cherry_picked}.
The plot only shows the directions and interactions needed to maintain 99.9\% accuracy on the task.
We can ablate most nodes (and even more interactions) while maintaining near-perfect
model performance.

This model is cherry-picked (out of 5 models trained with different seeds)
as our simplest modular addition model; LIB works particularly well on this model
and the interaction graph is easy to understand.
However we show quantitative results for all seeds in the following figures, and provide all interaction
graphs (including PCA versions) in Appendix \ref{appendix/modular_addition_all_seeds}.

\subsubsection{Functional relevance}
\label{subsubsec:mod_add_functional_relevance}
One advantage that LIB has over PCA is that it accounts for which features affect future layers and thus
can ignore computationally-irrelevant features. For instance, if a direction explains a lot of variance
in a layer's activations but doesn't affect future activations, then the PCA basis will include
it but LIB will not.

We test this manually by checking whether the LIB or PCA basis contain features that appear
irrelevant based in the interaction graph. We provide all interaction graphs and additional
details in Appendix \ref{appendix/modular_addition_all_seeds}. We find that in the final layer
the PCA has a lot of computationally-irrelevant features, while LIB does not. In the middle layer,
the effect is less strong, but we identify around 14 irrelevant features in the PCA basis and just
3 in the LIB basis (across the 5 seeds).

We conclude that when there are directions in the model's activation spaces that are not
computationally-relevant for the output but that explain a non-trivial fraction of the variance in a
layer, LIB is better than PCA at excluding these directions.

\subsubsection{Monosemanticity}
As a proxy for feature interpretability, we test how monosemantic LIB and
PCA features are. We measure monosemanticity as the fraction
of variance in the activation of a feature is explained by a single Fourier term.

\begin{figure}[htb]
    \centering
    \begin{subfigure}[b]{.72\textwidth}
        \centering
        \includegraphics[width=\textwidth]{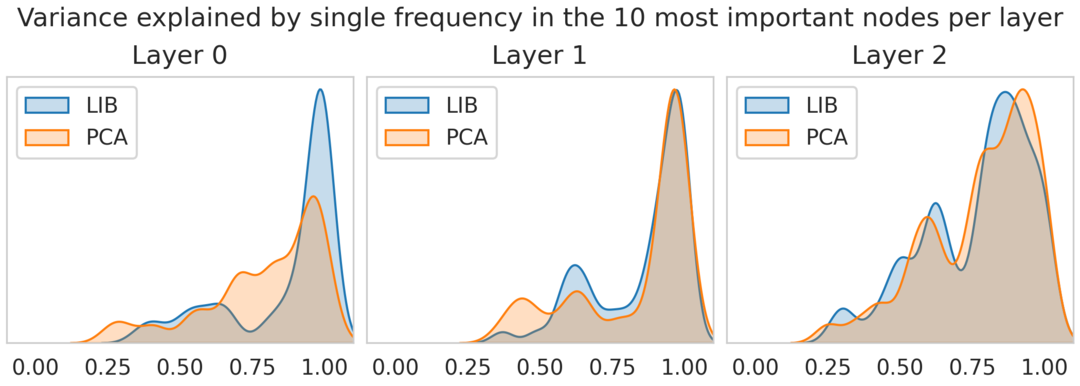}
        \caption{Variance explained by the first Fourier term (for all seeds, top 10 nodes per layer).}
        \label{fig:modular_addition_monosemanticity_kde}
    \end{subfigure}
    \hfill
    \begin{subfigure}[b]{.25\textwidth}
        \centering
        \begin{tabular}{ccc}
            \toprule
            Seed & LIB & PCA \\
            \midrule
            0 & 14 & \textbf{16} \\
            1 & 13 & 13 \\
            2 & 18 & 18 \\
            3 & 16 & 16 \\
            4 & \textbf{13} & 5 \\
            \bottomrule
        \end{tabular}
        \caption{Number of >90\% monosemantic features in the top 10 nodes of all layers, per seed.}
        \label{fig:modular_addition_monosemanticity_table}
    \end{subfigure}%
    \caption{Monosemanticity of features in LIB and PCA basis.}
    \label{fig:modular_addition_monosemanticity}
\end{figure}

We find that LIB does not have a consistent advantage over PCA.
Figure \ref{fig:modular_addition_monosemanticity_kde} shows the amount
of variance explained by the first Fourier term in the first 10
basis directions. We find no difference in layers 1 and 2, but an advantage
for LIB in layer 0. However, further investigation (Figure
\ref{fig:modular_addition_monosemanticity_table}) shows that
this is an outlier driven by seed-4 only.
Thus we conclude that LIB features are not clearly more monosemantic than PCA features.

\subsubsection{Sparsity}
\label{subsubsec:mod_add_sparsity}
The previous two sections analyzed the features of the LIB and PCA bases. Now we want to focus
on the \textit{interactions} between those features, as represented by the edges in the interaction
graphs. We start by comparing the sparsity of interactions in the two bases in this section, and
explore modularity in the next section.

\begin{figure}
    \centering
    \includegraphics[width=\linewidth]{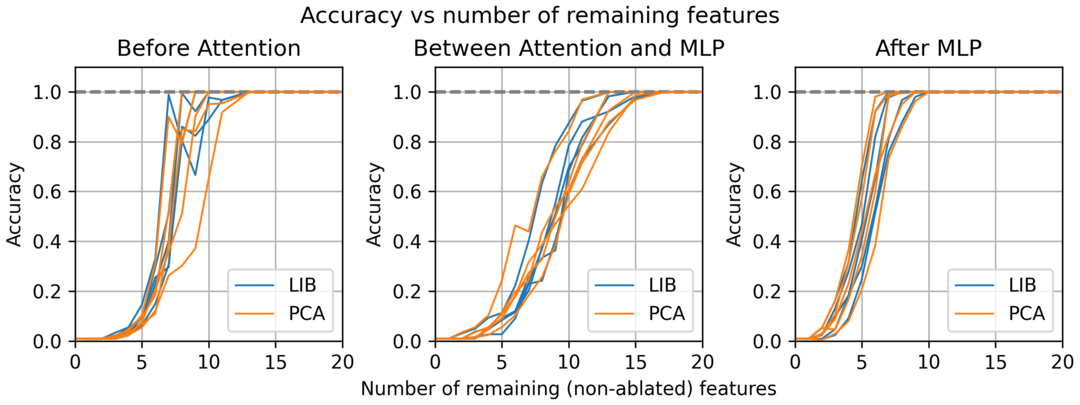}
    \caption{Number of nodes required to preserve >99.9\% accuracy for LIB and PCA on five
    modular addition transformers trained with different random seeds.}
    \label{fig:modular_arithmetic_nodes_required}
\end{figure}

To give some context on the interaction sparsity we briefly show the feature sparsity, i.e. the
number of features we can ablate without losing much accuracy. Figure
\ref{fig:modular_arithmetic_nodes_required} shows ablation curves, accuracy as a function of
remaining features (i.e. the left edge corresponds to all features being ablated). We show
LIB (blue) and PCA (orange) for 5 models trained with different seeds. LIB tends to require
slightly fewer features than PCA to maintain high accuracy, but the effect is very seed-dependent.

As our main metric for interaction sparsity we use edge ablations, as introduced
in Section \ref{method:edge_ablation}. Instead of ablating features, we ablate edges
and measure the accuracy as a function of the number of edges ablated. Rather than showing
the full edge ablation curves, we use the number of edges required to maintain
99.9\% accuracy as a benchmark in the modular addition task. Other thresholds yield
similar results.

\begin{figure}
    \centering
    \includegraphics[width=0.6\textwidth]{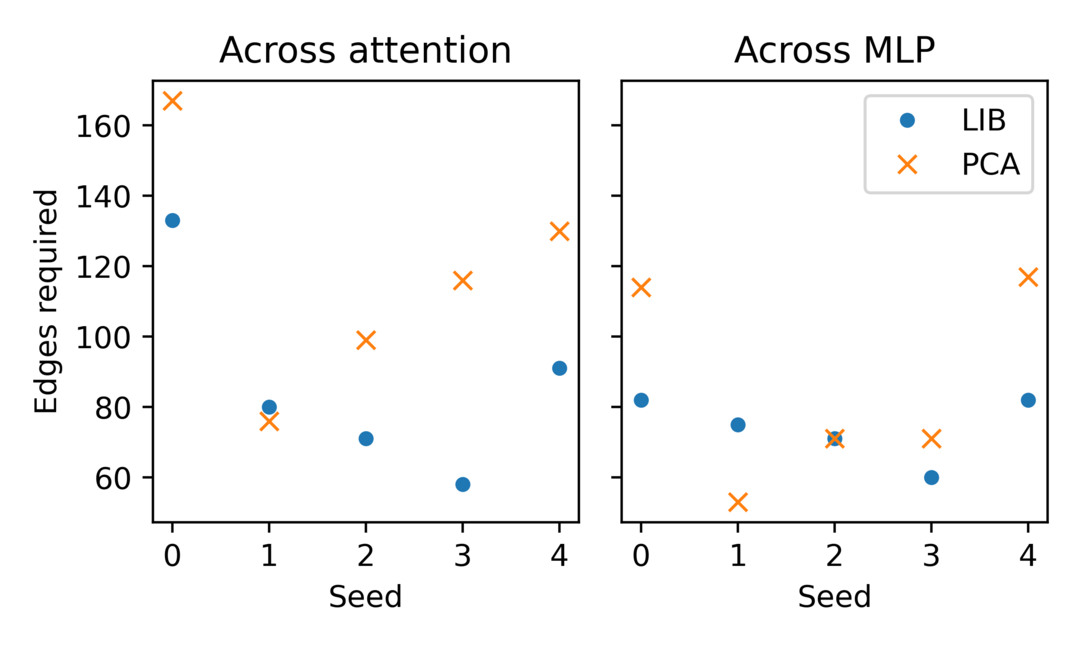}
    \caption{Number of edges required to preserve >99.9\% accuracy for LIB and PCA on five
    modular addition transformers trained with different random seeds.
    Lower number of edges required is better, as it means the representation has sparser interactions.
    Across attention LIB is always sparser than PCA. Across the MLP the trend is unclear.}
    \label{fig:modular_arithmetic_edges_required}
\end{figure}

We find mixed results, as shown in Figure \ref{fig:modular_arithmetic_edges_required}.
In the attention layer, the LI basis always requires fewer edges than the PCA basis, but
in the MLP layer LIB is sparser in only 3 out of 5 models, while PCA is sparser in 2 models.
We see the attention layer results as tentative evidence that LIB can find more sparsely interacting features than PCA,
but the MLP results are not clear-cut and depend on the model seed.

\subsubsection{Modularity}
Finally, we want to test whether our modularity algorithm as described in
Section \ref{subsec:modules} can find meaningful modules in the interaction graph. Ideally
we would want to see modules that cleanly separate the different frequencies used by the
model (which we expect to not interact much with each other).

We find that this works sometimes but not reliably. Figure \ref{fig:modular_arithmetic_rib_graph_cherry_picked}
shows a cherry-picked example (with fine-tuned resolution parameter $\gamma=0.5$, rather than the
default $\gamma=1$) where we find that
the modules mostly separate out the different frequencies used by the model (the most important
frequencies are 49, 38, and 36). Other models
(Appendix \ref{appendix/modular_addition_all_seeds}) show less clear-cut modules.
Overall, we conclude that the modularity algorithm we use does not achieve our goals reliably.

\subsection{CIFAR-10 MLP}
\label{subsec:cifar10}

As another simple test case, we apply LIB to small fully connected networks trained on CIFAR-10 \citep[][see Appendix \ref{appendix/models_cifar} for details]{krizhevsky2009learning}.
We train 5 different models with different random seeds with consistent
results; in this section we present results from seed-0 unless otherwise noted.

Our models achieve 46.6\% - 48.4\% test accuracy on CIFAR-10.
This is on par with other 2-layer fully connected MLPs without data augmentation, but only somewhat
better than pixel-based logistic regression (41\% accuracy) \citep{lin2015far}.
Basic CNNs can achieve much higher accuracy (e.g. \cite{krizhevsky2012conv} achieve 87\% accuracy), but adapting LIB to CNNs 
is out of scope for this paper.
Given its architecture, we expect our MLP model to use basic heuristics such as hue and brightness,
rather than relationships between neighboring pixels (which e.g. curve detectors require).

In this section, we compare the LI basis and PCA basis in two ways: We compare how sparsely features in different layers interact with each other (as in Section \ref{subsubsec:mod_add_sparsity}), and we analyze
how well both bases isolate a specific interpretable feature we
found, the vehicle-vs-animal feature.

\begin{figure}
    \centering
    \includegraphics[width=0.7\linewidth]{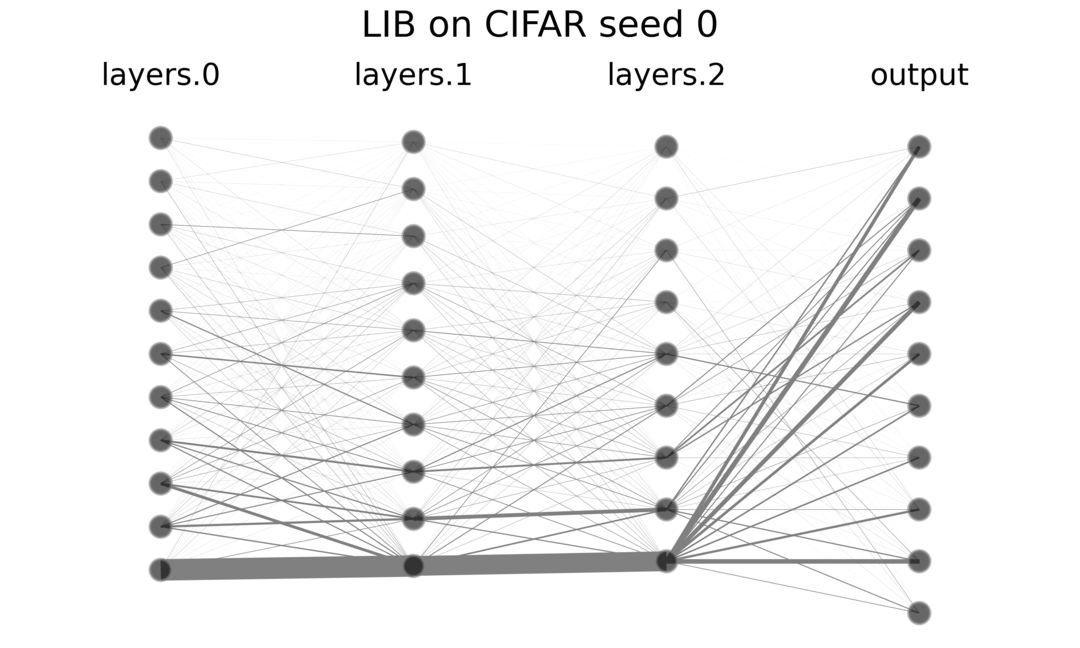}
    \caption{LIB interaction graph for our CIFAR model (seed-0). The four layers correspond to
    input, first hidden, second hidden, and output layer. Edges width corresponds to squared-edges, as this makes the graph more readable.}
    \label{fig:lib-graph-cifar}
\end{figure}

\subsubsection{Sparsity}
To judge the sparsity of interactions between the features in adjacent layers we
again use the edge ablation test (see Section \ref{method:edge_ablation}).
We measure how many interactions between features can be ablated while maintaining
a classification accuracy that is within 0.1 percentage points of the original model's accuracy.

\begin{figure}[h]
    \centering
    \includegraphics[width=0.4\textwidth]{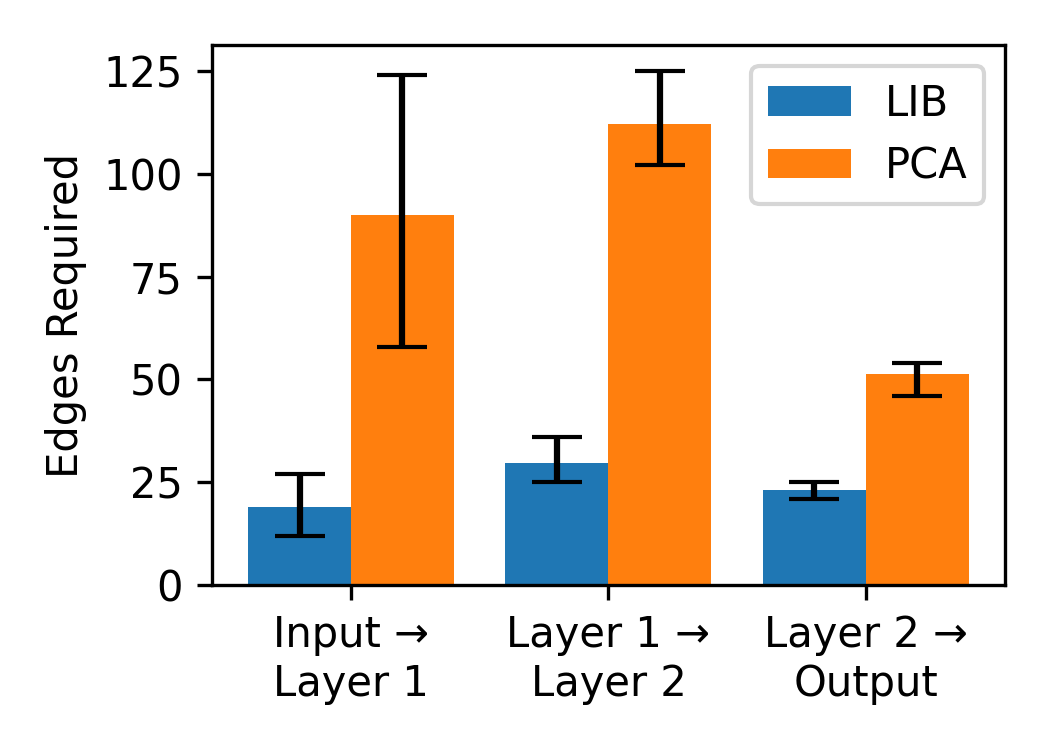}
    \caption{Edge ablation comparison between LIB and PCA on CIFAR-10. 
    The plot shows the minimum number of edges required to maintain
    an accuracy within 0.1 percentage points of the original model.
    Lower values indicate sparser interactions.
    The colored bars represent the mean across 5 random seeds, with
    the error bars indicating the minimum and maximum values across seeds.}
    \label{fig:cifar-sparsity}
\end{figure}

We find that, for all layers and seeds, interactions in the LI basis are sparser than in the PCA basis.
Figure \ref{fig:cifar-sparsity} shows the number of edges required, and we find that LIB always
requires fewer edges than the PCA basis.

Note that despite the somewhat dense looking graph in Figure \ref{fig:lib-graph-cifar}, we need very few edges (just ~25 connecting each pair of layers) to preserve good performance.

\subsubsection{Animal vs Vehicle Direction}
In Figure \ref{fig:lib-graph-cifar}, the LIB interaction graph, we see a path of thick edges at the bottom, indicating features which interact strongly with each other compared to other features in the network, and affect the output substantially. 
On inspection we notice that these features are all involved in distinguishing animals from vehicles. This is a core subtask on CIFAR-10 (containing 5 animal classes and 5 vehicle classes), and our model is reasonably good at it; summing the output probabilities across these categories gives a classifier with an AUROC of 0.946 (88.2\% accuracy).

LIB finds a single direction in the model at each layer that is almost solely responsible for this subtask, including in the `pixel space' in the input layer. This is an example of LIB successfully finding directions that are computationally-relevant for downstream parts of the model. If we use the direction in the final layer as a linear probe, it is nearly  as good of a classifier as the model itself, with an AUROC of 0.932.

\begin{figure}
    \centering
    \includegraphics[width=\textwidth]{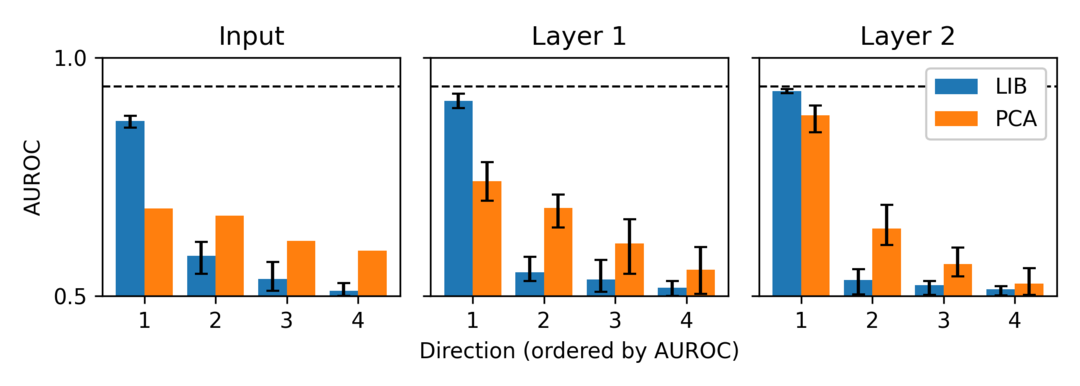}
    \caption{Comparison of how well LIB and PCA isolate the animal-vs-vehicle feature into a single
    basis direction.
    The four bars show the best animal-vs-vehicle classifying directions
    in each basis (measured by AUROC, again mean with errorbars indicating
    minimum and maximum values across 5 seeds). The dashed black line
    is the AUROC of the full model's output.
    LIB isolates the feature into a single direction, while PCA spreads it out over several directions.}
    \label{fig:cifar-animal-veh}
\end{figure}

In Figure \ref{fig:cifar-animal-veh} we show the AUROC of the top four LIB and PCA
features as animal/vehicle classifiers, to judge how well the animal/vehicle distinction
is concentrated to a single feature. We find that the animal-vs-feature direction
is clearly isolated in the LI basis, but not in the PCA basis. In the PCA basis
it is spread out over several directions.

We test whether our explanation of this direction as an animal-vs-vehicle feature is correct, we
perform two tests. First, we measure what fraction of variance in this
direction is explained by the animal-vs-vehicle label, and find that
the label only explains 54\% of the variance. This could be either because
the feature has a second function, or because the remaining variance is due to noise. To test this
we intervene on the feature, ``correcting'' the activation by setting
it to the mean value of activations with the true animal/vehicle label.
We find that this \textit{improves} the model, with accuracy increasing from 47.6\% to 54.6\%. This
is evidence that our interpretation of the feature as an animal-vs-vehicle feature is correct.

\subsection{Language Models}
\label{subsec:llms}

Finally, we apply the LIB method to language models, GPT2-small \citep{radford2019language}
and TinyStories-1M \citep{eldan2023tinystories}.
We describe model and dataset details in 
Appendix \ref{appendix/models_language}.
We focus on GPT2-small here because it is the
larger model. Our results for
TinyStories-1M are similar or better than for GPT2-small,
and shown in Appendix \ref{appendix/ablation-graphs}.

In this section, we show interpretations for some of the
LIB features (Section \ref{subsubsec:gpt2-interpretability})
and feature interactions (Section \ref{subsubsec:gpt2-interaction-graph}),
and test the sparsity and modularity of the graphs (Section \ref{subsubsec:gpt2-sparsity}).

\subsubsection{Interpretability of LIB directions}
\label{subsubsec:gpt2-interpretability}
We analyze a selection of LIB directions in GPT2-small
to understand if they track meaningful and interpretable features.

We find that we can successfully interpret the first 4 LIB directions,
which track positional features. We also show feature visualizations
for a random selection of other LIB directions, but find them to be
not particularly interpretable.

The results we find for the PCA basis are very similar to these LIB results and we omit them from
this section. This means the second rotation does not seem to aid interpretability, as we hoped it might.

\begin{figure}[ht]
    \centering
    \includegraphics[width=0.7\textwidth]{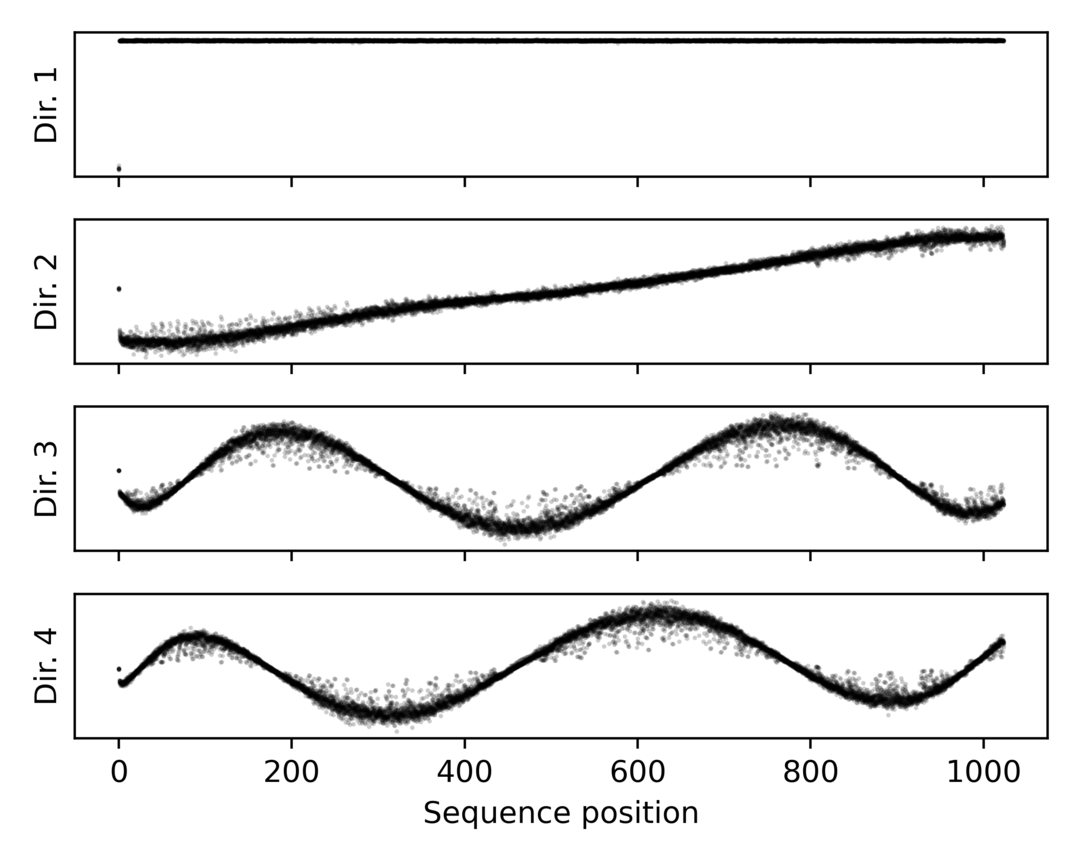}
    \caption{Positional features in GPT2-small. The activations of the first four LIB directions as a function of
    sequence index.}
    \label{fig:gpt2-pos}
\end{figure}

\paragraph{Positional features:}
We noticed that several directions in the LIB and PCA basis of GPT2-small seem to represent
positional information. We show this in Figure \ref{fig:gpt2-pos} where we plot the activations of
the first 4 LIB directions as a function of sequence position. We show the 6th block activations
as an example, but the plot is similar for throughout most layers of the model.
We find that around 97\% of the variance in these four directions is explained by the sequence position.
The first direction seems to represent a \enquote{is this
position 0?} feature, the second direction just scales with the sequence index, and the third and fourth
directions represent a sinusoidal function of the sequence index. The latter two directions
explain the helix-like structure in GPT2-small that has been noted before \citep[e.g.][]{yedidia2023helix}.

\paragraph{Language features:}
To visualize directions in activation space we show dataset examples that activate the feature
by a given amount. Our visualization is based on \texttt{sae\_vis} \citep{sae_vis}, but modified
because our features, unlike features from standard sparse autoencoders, can be both positive and negative. We show the minimum and
maximum activating dataset examples, as well as four intervals of intermediate-activating examples.
We again show features in the 6th block of GPT2-small (specifically at the very beginning of the
block) but the results for other blocks are qualitatively similar.
We host feature visualizations for a random selection of LIB and PCA features of all layers at
\url{https://data.apolloresearch.ai/lib/feature_viz/}. Below we provide screenshots
for a selection of LIB directions.

\begin{figure}[ht]
    \centering
    \adjincludegraphics[width=\textwidth,trim=0 0 0 0.15\height,clip]{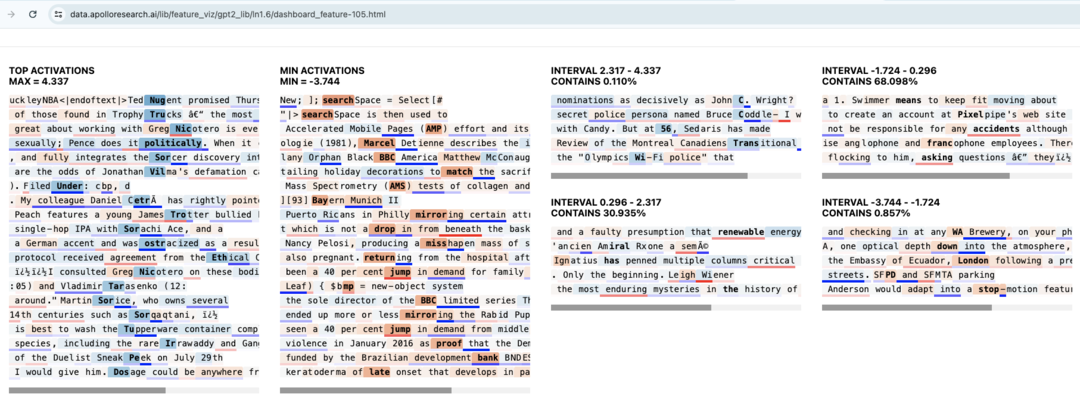}
    \caption{LIB feature 105 at the beginning of the 6th block of GPT2-small.}
    \label{fig:gpt2-feature-pos}
\end{figure}

As an example of a somewhat interpretable direction we show feature 105 in Figure
\ref{fig:gpt2-feature-pos}. The feature seems to maximally activate on the first
token of last names and brand names (``Ted Nugent'', ``Greg Nicotero'', ``Tupperware''),
and we also see two such examples in the first quartile samples (``John C. Wright'', ``Bruce
Coddle''). However, we also see other samples that do not follow this rule.

\begin{figure}[ht]
    \centering
    \adjincludegraphics[width=\textwidth,trim=0 0 0 0.15\height,clip]{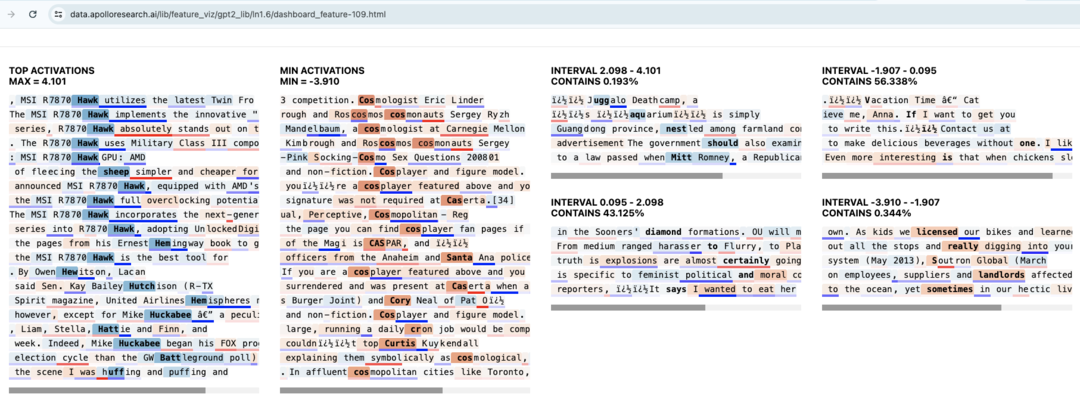}
    \caption{LIB feature 109 at the beginning of the 6th block of GPT2-small.}
    \label{fig:gpt2-feature-neg}
\end{figure}

One worry we have is that the maximum and minimum activating examples are dominated by outliers
in the dataset, and not indicative of the meaning of the direction.
As an example of this we show feature 109 in Figure \ref{fig:gpt2-feature-neg}. Most of the
top activating examples are the word ``Hawk'' in the name of a particular graphics card, and other top activations
are last names starting with H. Minimally activating examples are mostly the token ``cos'' and
similar tokens. However, none of these interpretations explain the dataset examples that cause intermediate activations,
showing that the interpretation based on top-activating samples is not predictive and likely not correct. Possible ways to improve this
include computing feature activations on custom prompts designed to test an interpretation, and
using more intermediate samples. A detailed interpretability analysis however is out of scope for
this paper.
In Appendix \ref{appendix/llm_features} we show more visualizations of three more randomly selected
LIB features.

Overall we find that the LIB features are not particularly interpretable. This is comparable to
what we find for PCA directions (not shown here but available at this
\href{https://data.apolloresearch.ai/lib/feature_viz/}{url}).

\subsubsection{Interpretability of interactions}
\label{subsubsec:gpt2-interaction-graph}

\begin{figure}
    \centering
    \includegraphics[width=1\linewidth]{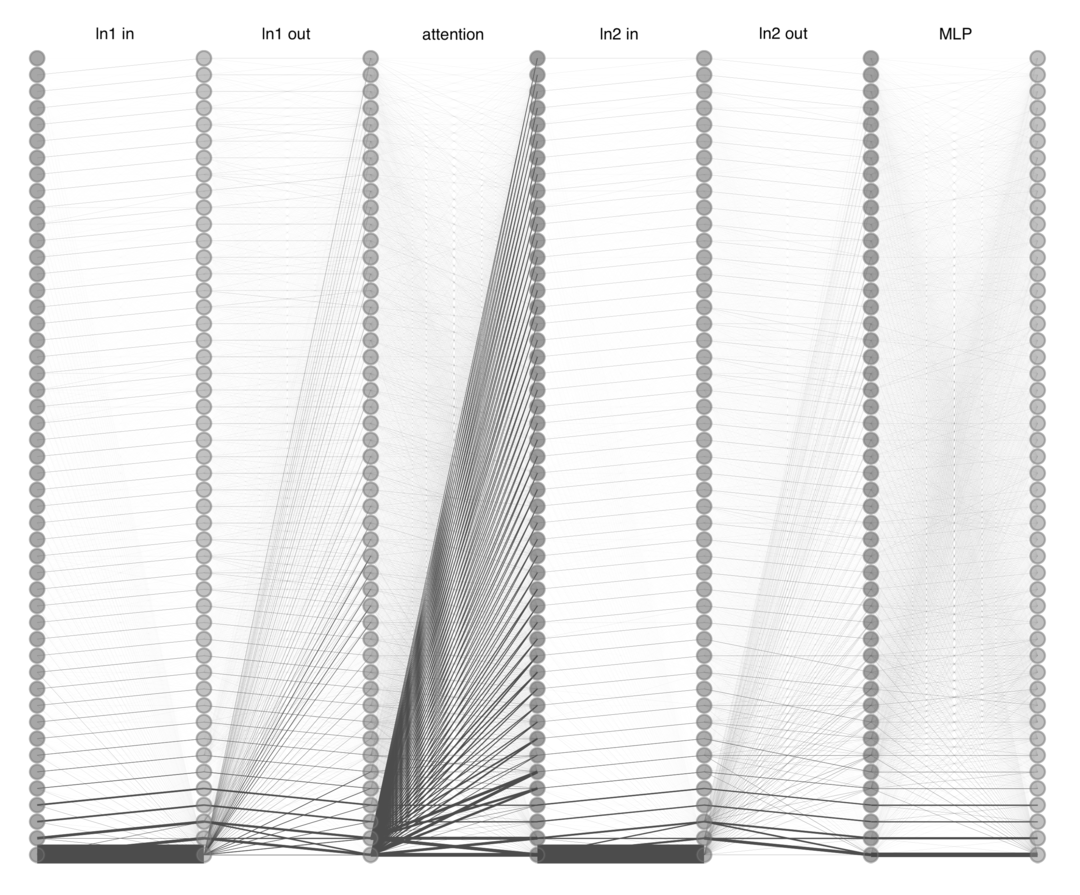}
    \caption{The LIB interaction graph at a representative block in the middle of GPT2-small (block 6). All edges have been square-root normalized. Only the first 50 nodes in each layer are shown. The size of the two largest edges have been clipped to make the graph readable.}
    \label{fig:gpt2-graph-L6}
\end{figure}

We use the interaction graph to analyze the interactions between LIB features in GPT2-small. We show
one transformer block of the graph in Figure \ref{fig:gpt2-graph-L6}. We have arbitrarily
chosen transformer block 6, but the other blocks look similar.
Out of the six sets of edges shown, two correspond to attention and
MLP layers, while the remaining four are related to layer norm (two layer norms, each split into two sets of edges).
We provide a diagram of our transformer architecture in Appendix \ref{appendix/sequential_transformer}.

We find that the edges corresponding to layer norms are mostly 1-to-1 connections. The only exceptions are the
connections to and from the variance feature.
The variance feature is influenced by all input features, but most strongly by the first LIB feature
(edges labeled ``ln1 in'' and ``ln2 in'' in Figure \ref{fig:gpt2-graph-L6}).
As we saw in Section \ref{subsubsec:gpt2-interpretability}, this feature is a ``is this the 0th
position in the sequence''-feature. This explains its outsized effect on the variance feature
because the 0th position is treated uniquely in GPT2 and has an unusually large norm. The variance
feature is also connected to all output features (edges labeled ``ln1 out'' and ``ln2 out'' in Figure \ref{fig:gpt2-graph-L6}).
This is expected because the variance in layer norm scales the output in every direction.

Looking at the edges across the attention module, we see the first two input features have a strong connection to most of the output features. These input features are a binary representation of ``is this position 0?'' and a linear representation of ``sequence index'' respectfully (see Section \ref{subsubsec:gpt2-interpretability}).
The LIB graph claims that these input features are influential on most output features of the attention block.
This is expected because we know that the attention mechanism often depends on the token position.

Apart from these two observations about positional information, we cannot make much
sense of the interaction graph. In particular the edges across Attention and
MLP layers are relatively dense and did not permit straightforward interpretation with manual inspection.

\subsubsection{Sparsity}
\label{subsubsec:gpt2-sparsity}
We run our edge-ablation test again (see Section \ref{method:edge_ablation}) to measure
how many edges we can ablate while maintaining a cross-entropy loss within 0.1
of the original model. This is a relatively large loss increase; we do not claim that a 0.1
loss increase is inessential but merely use it as an (arbitrary) threshold to compare
the sparsity of the LIB and PCA bases. 

\begin{figure}
    \centering
    \includegraphics[width=0.8\linewidth]{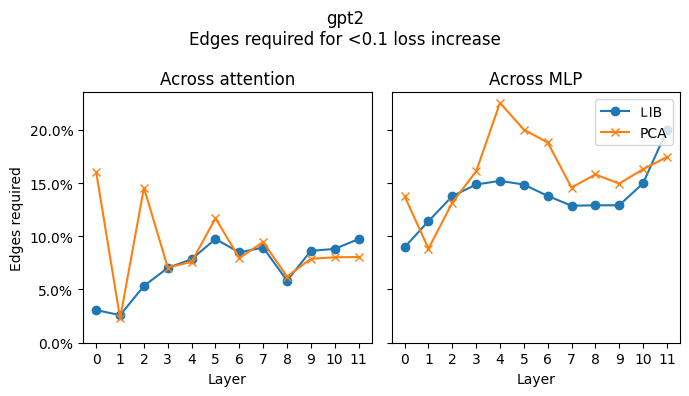}
    \caption{Edge ablation results on GPT2-small, for both LIB and PCA interaction graphs. We ablate as many edges as possible without increasing the cross-entropy loss by more than 0.1. The fewer edges, the better since this implies a sparser interaction graph.}
    \label{fig:gpt2-eabl-attn-mlp}
\end{figure}

We show the results in Figure \ref{fig:gpt2-eabl-attn-mlp}. We find that, in both bases, we can ablate
80 to 95\% of the interactions while keeping the loss increase below 0.1.
We can ablate slightly more interactions using LIB compared to PCA but the results are noisy.
LIB tends to give sparser interactions than PCA in some
early attention layers and in most MLP layers.

In Appendix \ref{appendix/ablation-graphs} we run the same test for the much-smaller TinyStories-1M
model and find that LIB is consistently sparser than PCA in the attention and MLP layers of TinyStories-1M,
as shown in Figure \ref{fig:tinystories-edge-abl-all}. For the TinyStories-1M model we were able to use
a larger dataset to compute the LIB and PCA bases. These results suggest that either LIB works better
on the smaller model, or that the larger dataset improved the quality of the LI basis. We do not
test LIB on GPT2-small with a larger dataset because even if we achieved
sparsity improvements similar to Figure \ref{fig:tinystories-edge-abl-all}, say increasing the number
of interactions we can ablate from $\sim 80\%$ to $\sim 90\%$, we would still not consider this result good enough to achieve our goals. LIB would very likely not enable qualitatively different interpretability work or
fine-grained modularity analysis on GPT2-small.

For completeness, we also show results for non-attention and non-MLP layers in appendix
\ref{appendix/ablation-graphs}. The results for those layers are mixed, with very different
results between GPT2-small and TinyStories-1M.
However, we think that results for those layers are less relevant, as
they just represent the layer norm layers and are very sparse in both PCA and LIB basis.

%% file: sections/conclusions.tex
\section{Conclusion}
\label{sec:conclusion}
We developed a novel interpretability method based on a transformation to the local interaction basis (LIB)
and integrated gradient interaction graphs.
Our work is built on the theoretical work
by \citet{apolloTheory} who propose representations based on parameterization-invariant structures in neural networks.
LIB assumes that features in a neural network
can be represented in a non-overcomplete basis. Additionally we assume that the main sources of freedom in the loss-landscape
are linear dependencies in the activations of individual layers, and linear dependencies in the gradients between adjacent
layers. LIB attempts to find a basis of sparsely interacting features for the activations of a neural network, represent the interactions between features in a graph, and identify modularity in the network's computations by searching for modules in this graph.

We find that our method produces more interpretable and more sparsely-interacting representations on toy models (a modular
addition transformer and a CIFAR-10 model), compared to a baseline of using PCA directions as
features.
On language models, we find that LIB produces more sparsely-interacting representations than
PCA in some layers, but the representations are not more interpretable. Furthermore, we do not find evidence of
modularity in the interaction graphs produced by LIB on language models.

Our goal was to test whether LIB could find a basis of features that would be more interpretable and
interact much more sparsely than baseline methods (PCA). We conclude that, while promising on toy models,
LIB does not achieve this goal on language models.
We speculate that the assumption of a linear non-overcomplete basis was wrong in the case of LMs,
and is the reason for not finding interpretable and sparsely-interacting features.

While our test of the LIB method yielded a negative result, we
are still excited about future work building
other methods based on the theory-framework \citep{apolloTheory}. %
In particular, we are interested in developing a generalization of LIB to the case of overcomplete bases to allow for the possibility of the network representing features in superposition using sparse coding.

\section{Contribution Statement}
\label{sec:contribution}

\textbf{Theory}
\LB led the theory development, working on the conceptual framework, motivation and LIB methodology
before anyone else joined the project. \JM provided extensive red-teaming and development of the
methodology via theoretical arguments and experimental counterexamples. \KH first proposed the use
of integrated gradients and contributed substantially to refinements of the theory. 

\textbf{Infrastructure}
\DB led development of the current codebase, with significant contributions from \NGD and \SH. This replaced much earlier versions by \JS, \AG and \MH. \NGD and \DB scaled
the implementation to LLMs. \NGD implemented the edge ablation experiments. \SH implemented most of
the supported bases and interaction metrics. The main experiments were run by \SH and \NGD.

\textbf{Analysis}
Much of the analysis of earlier versions of LIB was performed by \MH on a modular addition
transformer, an MNIST MLP, and GPT2-small, with support from \AG and \JS. Of the analysis that this
manuscript is based on, \SH led the analysis of the Modular Addition Transformer and
\NGD led the analysis of the CIFAR10 model. \SH and \NGD contributed equally to the analysis of the
language models.
 
\textbf{Paper Writing}
The manuscript was primarily written by \SH. Sections
\ref{section/intro} and \ref{section/methodology} were
drafted by \LB and \JM with extensive contributions by \MH.
The experiment section was drafted by \SH (modular addition and LLM feature visualization) and \NGD (CIFAR and LLM positional features). Diagrams were created by \JM and \SH.

\textbf{Leadership}
\LB coordinated the team, leading the high-level direction around which experiments to run.
\SH coordinated the writing of this manuscript.
\MH acted as a hands-on advisor, managing the team and giving regular feedback throughout the
project and the writing of this manuscript.

\section{Acknowldgements}
We are grateful to James Fox, Jacob Hilton, Tom McGrath, and Lawrence Chan for feedback on this manuscript. Apollo Research is a fiscally sponsored project of Rethink Priorities.

%% file: sections/appendix.tex
\section{Models and datasets}
\label{appendix/models}
Here we describe the models and datasets we use in the modular addition, CIFAR-10, and language model
experiments. We also describe our slightly modified sequential transformer architecture.

\subsection{Modular addition}
\label{appendix/models_modular}
Our modular addition model follows \citet{nanda2023progress}. It is a 1-layer decoder-only
transformer with a residual stream width of 128, 4 attention heads of width 32, an
MLP block of width 512 with ReLU activation, and no layer norm.

The dataset consists of sequences [$x$, $y$, $=$] where $x$ and $y$ are numbers from 0 to 112,
and $=$ is a constant. The labels (tested at the third position) are $z = (x + y) \mod 113$;
we don't evaluate or compute the output at the first and second positions, as they do not affect the loss.
The total dataset has 113*113=12769 sequences; we use the first 30\% to train the model and PCA/LIB,
and the rest for testing.

We train the models with learning rate 0.001 (linear schedule with warmup),
batch size 10,000, and the AdamW optimizer ($\gamma=1, \beta_1=0.9, \beta_2=0.98$).
We train 5 models with different random seeds. The models are trained for 60,000 epochs, and we
observe \enquote{grokking}, i.e. a sharp decrease in test loss, after about a third of that time.
All models achieve 100\% accuracy on the train and test sets.

\subsection{CIFAR-10}
\label{appendix/models_cifar}
Our CIFAR-10 \citep{krizhevsky2009learning} models are feedforward MLPs with two hidden layers. Each hidden layer has 60 neurons with a ReLU activation. We train 5 models that achieve
46.6\% - 48.4\% accuracy on the test set.

\subsection{Language models}
\label{appendix/models_language}
For the language model tests we use pre-trained transformers. In this paper we show results for
GPT2-small \citep{radford2019language} and Tinystories-1M \citep{eldan2023tinystories}. Both
models follow the GPT2 architecture, including layer norm.
GPT2-small is the larger model with 85M parameters, 12 layers, and a residual stream width of 768.
Tinystories-1M (1M parameters) has 8 layers and a residual stream width of 64.

We use the following datasets to run the LIB method:
For GPT2 we use the openwebtext dataset \citep{Gokaslan2019OpenWeb}, specifically a pre-tokenized
version we host at \texttt{apollo-research/Skylion007-openwebtext-tokenizer-gpt2}.
We run PCA on 50,000 sequences (51M tokens), and the second transformation of LIB (see Section \ref{subsec:LIB}) on Jacobians collected over 500 sequences (512,000 tokens),
which takes on the order of 24 GPU-hours on A100 GPUs.

For Tinystories-1M we use the Tinystories dataset, again we host a pre-tokenized version
(\texttt{apollo-research/skeskinen-TinyStories-hf-tokenizer-gpt2}). We use 99\% of the dataset for
the PCA, and 50,000 sequences (10M tokens) for the second transformation of LIB which is the
limiting factor in terms of computational cost. We use the last 1\% of the data as a test set.

\subsection{Sequential transformer}
\label{appendix/sequential_transformer}
The LIB method assumes models to be sequential, i.e. each layer can be written as a function
of the previous layer's output. To accommodate this, we make slight modifications to the transformer
architecture that do not affect the computation, but simply the way we write the model.
Essentially we make the residual stream an input to every layer that is just passed through
to its output.

\begin{figure}
    \centering
    \includegraphics[width=\textwidth]{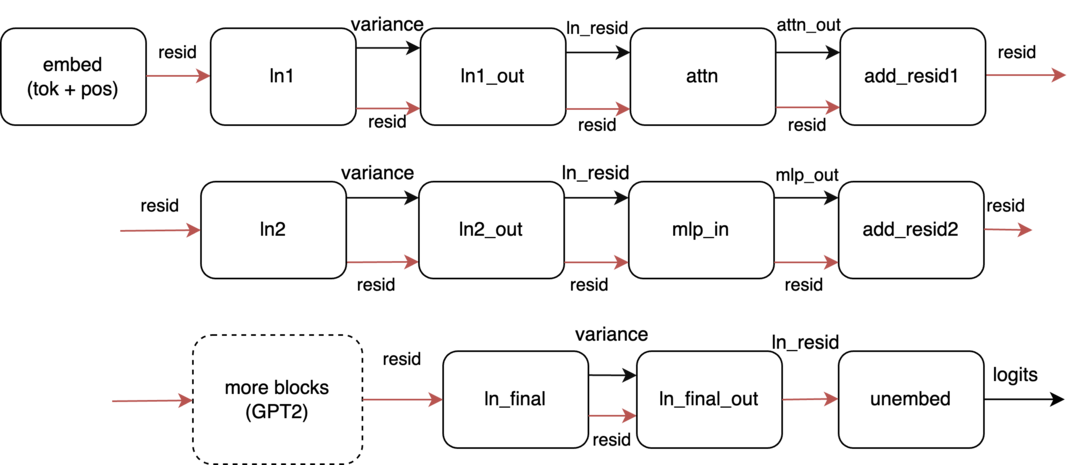}
    \caption{Sequential transformer.}
    \label{fig:sequential_transformer}
\end{figure}

We show the resulting setup in Figure \ref{fig:sequential_transformer}. The diagram shows
a transformer block starting the the features before ``ln in'' (named \texttt{ln1}),
the features between ``ln in'' and ``ln out'' (named \texttt{ln1\_out}), and the features after
``ln out'' before the attention layer (named \texttt{attn\_in}). The next three feature layers
follow a similar pattern with \texttt{ln2}, \texttt{ln2\_out}, and \texttt{mlp\_in}.
This is the naming convention we use in the feature visualizations hosted at
\href{https://data.apolloresearch.ai/lib/feature_viz/}{this url}.

\subsection*{Layer norm treatment}
Layer norm is a reasonably simple operation that (if handled naively) often appears messy and
convoluted in LIB graphs. This is because every input feature affects the variance in the
denominator of layer norm, and this denominator affects all output features.
This all-to-all interaction can make it harder to disentangle modules.

In order to avoid this issue we split the layer norm module into two sections, where the
``variance" used by layer norm is a hardcoded to be its own feature, and excluded from the
LIB transformations.

The first part of layer norm (``ln in'') just computes the variance and concatenates it to the
residual stream. The second part of layer norm (``ln out'') normalizes all other features
by dividing by this variance. Splitting layer norm into two steps creates two simple layers
rather than one complex layer, making the interaction graph more interpretable.

\subsection{Folding-in biases}
\label{appendix/bias_fold}

It is mathematically convenient to consider a network without bias terms: this
makes attribution easier as zero input results in zero output, and simplifies
the mean centering by making it a linear operation.

In MLP and Transformer models, biases appear in the form
$\mathbf{f}^{l+1} = \text{act}(W^l \mathbf{f}^l + \mathbf{b}^l)$,
with the weight matrix $W^l$, the bias vector $\mathbf{b}^l$, and
the activation function $\text{act}$. Throughout this paper
we redefine the activations and weights as
\begin{align}
    \mathbf{f}^{l} = \begin{pmatrix} 1 \\ \mathbf{f}^{l}_\text{orig} \end{pmatrix}
    \in \mathbb{R}^{d^l+1}
    \quad\text{and}\quad
    W^l = \begin{pmatrix} 1 & 0 \\ \mathbf{b}^l_\text{orig} & {W}^l_\text{orig} \end{pmatrix}
    \in \mathbb{R}^{(d^{l+1}+1) \times (d^l+1)}
\end{align}
so that we can write the network without any biases as $\mathbf{f}^{l+1} = \text{act}(W^l \mathbf{f}^l)$.

\section{Mathematical description of LIB transformations}
\label{appendix/LIB_transformations}
We describe the LIB transformation as a PCA of the activations, and an SVD of the Jacobians.
In this sections we provide a more detailed explanation of the LIB transformation, and equations
to reflect our implementation accurately. We also provide pseudocode for the LIB and IG code in
Appendix \ref{appendix/pseudocode}.

We express the LIB linear transformation as a matrix, $\hat{\mathbf{f}}^l = C^l \mathbf{f}^l$,
where $\mathbf{f}^l$ is the vector of activations in layer $l$ and $\hat{\mathbf{f}}^l$ is the
vector of activations in the LIB basis. We can factor this transformation into the two steps,
$C^l = C^l_{\rm SVD} C^l_{\rm PCA}$. In the rest of this section we will derive how these
transformations are composed until we arrive at the final equation for $C^l$
\begin{gather}
    C_{\rm PCA}^l = ({D^{l}}^\frac{1}{2})^+ U^{l} \centermatrix^{l}\,, \quad
    C_{\rm SVD}^l = {\Lambda^{l}}^{\frac{1}{2}} V^{l}\ \\
    C^{l}={\Lambda^{l}}^{\frac{1}{2}} V^{l} ({D^{l}}^\frac{1}{2})^+ U^{l} \centermatrix^{l}\,.
\end{gather}
We will explain the individual components in the following sections.

\subsection{The first transformation}
\label{appendix/first_transformation}
The first transformation implements a PCA of the activations in layer $l$. This consists
of three steps: centering the activations, transforming them into a basis aligned with the
principle components of the activations over the data set, and rescaling the basis directions to
have a variance of 1.

Starting with the first step, we define the matrix $\centermatrix^{l}$ 
that centers the activations over the (training) dataset as
\begin{align}
\centermatrix^{l}_{jj'} = \delta_{jj'}-\mathbb{E}(f^{l}_j) \,\delta_{0j'}
\end{align}
We can write this as a single matrix
because the activations include a constant column due to our bias-folding (appendix
\ref{appendix/bias_fold}).

Next we derive the principal components by solving the eigenvalue problem for the Gram matrix
$G^{l}$ of the centered activations. We obtain the diagonal eigenvalue matrix $D^{l}$ and the
orthogonal projection matrix $U^{l}$:
\begin{gather}
G^{l}_{jj'}\coloneq \frac{1}{\vert \dataset\vert} \sum_{x \in \dataset} {\left(\centermatrix^{l} \mathbf{f}^{l}(x)\right)}_j {\left(\centermatrix^{l} \mathbf{f}^{l}(x)\right)}_{j'}\\
G^l \eqcolon {U^{l}}^T D^{l} U^{l}
\end{gather}

Finally we use a multiplication by $({D^{l}}^\frac{1}{2})^+$ to rescale the activations, and to
remove near-zero PCA components. The $+$ denotes the Moore-Penrose pseudo-inverse of
${D^{l}}^\frac{1}{2}$. In the case of a diagonal matrix this gives another diagonal matrix with
diagonal entries given by the reciprocal of the diagonal entries in the original matrix -- unless
the original entry was zero in which case the new entry is also zero. In our implementation of the
pseudo-inverse, some principal components that are extremely close to zero ($<10^{-15}$ by default)
are treated as exactly zero.

We obtain the intermediate result
\begin{gather}
    \tilde {\mathbf f}^l = ({D^{l}}^\frac{1}{2})^+ U^{l} H^l \mathbf{f}^l\,.
\end{gather}

\subsection{The second transformation}
\label{appendix/second_transformation}
The second transformation attempts to sparsify the interactions by aligning the basis of the
activations in layer $l$ with the basis in layer $l+1$. We do this by performing an SVD of the
Jacobians, the derivative of the activations in layer $l$ with respect to the activations in layer
$l+1$ for all data points.

To do this, we first compute the Jacobians
\begin{gather}
    J^{l}_{ij}(x) \coloneq \frac
            {\partial\hat{f}^{{l+1}}_i({\tilde{\mathbf{f}}}^{l}(x))}
            {\partial {\tilde{f}_j}^{l}(x)} %
\end{gather}
obtaining a 3-tensor with dimensions $(\vert \dataset \vert, d^{l+1}, d^l)$. We flatten the first
two indices and calculate the right-handed SVD (via eigendecomposing the gram matrix $M$).
\begin{gather}
\label{eq:M} {M}^{l}_{jj'}\coloneq 
    \frac{1}{\vert \dataset \vert}
    \sum_{x \in \dataset}\sum_{i=0}^{d^{l+1}}
        J_{ij}^{l}(x) J_{ij'}^{l}(x)
        \\
        \label{eq:eigendecomp_M}
        M^{l} \eqcolon {V^{l}}^T \Lambda^{l} V^{l}\,.
\end{gather}
In the eigendecomposition \eqref{eq:eigendecomp_M} we exclude the $j=0$ constant direction
(responsible for mean centering) which we want to keep isolated and unchanged.

We obtain the orthogonal matrix $V^{l}$ which transforms the activations into
a basis aligned with the (right) singular vectors of the Jacobian, and the diagonal matrix
${\Lambda^{l}}^{\frac{1}{2}}$ which rescales the activations by how important they
are for the next layer. We obtain the final activations as
\begin{align}
    \hat{\mathbf{f}}^l &= V^{l} ({\Lambda^{l}}^{\frac{1}{2}}) \tilde{\mathbf{f}}^l
    \\
    &= V^{l} ({\Lambda^{l}}^{\frac{1}{2}}) ({D^{l}}^\frac{1}{2})^+ U^{l} H^l \mathbf{f}^l\,.
\end{align}

\subsection{Recursive procedure}
Since the second basis transformation depends on the basis chosen for the following layer, we
calculate the LI basis recursively, starting from the final layer $l_{\text{final}}$ and working
backwards. At the final layer, we initiate the recursion with the PCA basis (without rescaling)
\begin{equation}
\begin{aligned}\label{eq:recursion_start_eqns}
&C^{l_\text{final}}\coloneq U^{l_\text{final}} H^{l_\text{final}}.
\end{aligned}
\end{equation}

\subsection{Transformer implementation:}
\label{appendix/transformer_implementation}
For transformers, the activations in a layer are a sequence of activations over the token dimension.
In the first transformation we can simply treat the token dimension $t$ just like the data
dimension $x$, expanding the sum to
\begin{equation}
    G^{l}_{jj'}\coloneq 
\frac{1}{T \vert \dataset \vert}
\sum_{x \in \dataset}\sum_{t=1}^T 
{\left(\centermatrix^{l} f^{l}(x)\right)}_{j,t}
{\left(\centermatrix^{l} f^{l}(x)\right)}_{j',t}\,.
\end{equation}
For the second transformation we need to take into account that activations in adjacent layers at
different token positions can depend on each other, so we need to compute the gradients for
all combinations of token positions $s,t$. We again flatten the now 5-dimensional ($x,s,i,t,j$)
Jacobian along all dimensions except for $j$ to compute the right-handed (``$j$-sided'') SVD:
\begin{equation}\label{eq:G_M_transformer}
M^{l}_{jj'}\coloneq \frac{1}{T \vert \dataset \vert}
\sum_{x \in \dataset}
\sum_{s,t=1}^T
\sum_{i = 1}^{d^{l+1}} \frac{\partial \hat{f}^{{l+1}}_{i,s}({\tilde{\mathbf{f}}}^{l}(x))}{\partial \tilde{f}^{l}_{j,t}(x)}\frac{\partial \hat{f}^{{l+1}}_{i,s}({\tilde{\mathbf{f}}}^{l}(x))}{\partial \tilde{f}^{l}_{j',t}(x) }\,.
\end{equation}

Transformers are typically represented as a residual stream with MLPs and attention components
in parallel. For our method we want feature layers to represent a causal cut through the network,
so we concatenate the activations of parallel components (in the standard basis) into a single
vector $\mathbf{f}^{l}$ before the LIB transformation. We illustrate this \enquote{sequential
transformer} architecture in Appendix \ref{appendix/sequential_transformer}.

\section{Stochastic sources}
\label{appendix/stoch_sources}
Calculating the LIB basis, as well as the integrated gradient attributions, involves computing
Jacobians between adjacent layers in the network (equations \ref{eq:edges_transformers} and
\ref{eq:G_M_transformer}). This can be computationally expensive, especially for transformers,
since their Jacobians have two hidden and two position indices and thus require calculating many
entries.

\subsection{Stochastic sources in integrated gradient attribution}

To make the calculation cheaper, we use stochastic source techniques (see e.g. chapter 3.6 in
\citet{knechtli2016lattice} for an introduction). The idea is to not calculate the full Jacobian
for each data point, but instead to calculate the gradient for a few random directions instead.
We now demonstrate the derivation for the attribution calculation, the basis calculation is analogous.
Equation \eqref{eq:edges_transformers} can be written as
\begin{align}
    {\left(E^{l+1,l}_{i,j}\right)}^2 &= \frac{1}{\vert \dataset \vert}\frac{1}{T}\sum_{x\in \dataset}\sum^T_{s=1}
    A^{l+1,l}_{(i,s),j}(x)A^{l+1,l}_{(i,s),j}(x) \label{eq:edges_transformers_withA}
    \\
    &\text{with} \ A^{l+1,l}_{(i,s),j}(x) = \sum^T_{t=1}f^{l}_{j,t}(x)\int^1_{0}\text{d}\alpha \left[\frac{\partial}{\partial z^l_{j,t}} \left(F^{l+1,l}_{i,s}(\mathbf{z}^l)\right)\right]_{\mathbf{z}^l=\alpha \mathbf{f}^{l}(x)}\,.
\end{align}
We are computing all entries of the $\mathbf{A}^{l+1,l}_{(i,s)}$ vector, but we do not care about
the individual entries, only the sum of squares. This is a situation where stochastic sources
can be applied. Intuitively this means, instead of calculating every entry of
$\mathbf{A}^{l+1,l}_{(i,s)}$, we calculate the projection of this vector into a couple of random
directions and sum these up instead.

Mathematically we express this as inserting an identity into equation \eqref{eq:edges_transformers_withA}
and replacing it with $ \delta_{s,s'} \approx \nicefrac{1}{R}\sum^R_{r=1} \phi_{r,s} \phi_{r,s'}$
where the sources $\mathbf{\phi}_{r}$ are random directions\footnote{Specifically we use
$\phi_{r,s}\in \{+1,-1\}$ with equal probability. This has been shown to be optimal if we use
no additional information \citep{Dong_1994}.}, independently drawn for every $r$ and every sample $x$
from a distribution with mean zero and variance one
\begin{align}
    {\left(E^{l+1,l}_{i,j}\right)}^2 &= \frac{1}{\vert \dataset \vert}\frac{1}{T}\sum_{x\in \dataset}\sum^T_{s=1}\sum^T_{s'=1} \delta_{s,s'} A^{l+1,l}_{(i,s),j}(x)A^{l+1,l}_{(i,s'),j}(x)\\
     &\approx \frac1R\frac{1}{\vert \dataset \vert}\frac{1}{T}\sum_{x\in \dataset}\sum^R_{r=1} \left(\sum^T_{s=1} \phi_{r,s}(x) A^{l+1,l}_{(i,s),j}(x)\right) \left(\sum^T_{s'=1}\phi_{r,s'}(x) A^{l+1,l}_{(i,s'),j}(x)\right)\,.
    \label{eq:edges_transformers_stoch}
\end{align}
The terms in brackets require the same number of gradient calculations as the original
$A^{l+1,l}_{(i,s),j}$ term
\begin{gather}
    \left(\sum^T_{s=1} \phi_{r,s}(x) A^{l+1,l}_{(i,s),j}(x)\right) =  \sum_{t=1}^T f_{j,t}^l(x)
    \int_0^1  \text{d}\alpha \left[\frac{\partial}{\partial z_{j,t}^l} \left(\sum_{s=1}^T \phi_{r,s}(x) F^{l+1,l}_{i,s}(\mathbf{z}^l)\right)\right]_{\mathbf{z}^l=\alpha \mathbf{f}^l(x)}\,.
\end{gather}
Thus using equation \eqref{eq:edges_transformers_stoch} over \eqref{eq:edges_transformers} changes
the computational cost from $\mathcal{O}(T^2)$ to $\mathcal{O}(RT)$. This approximation is exact 
only in the limit $R \rightarrow \infty$, but typically is good enough for $R<T$ and thus reduces
the computational cost significantly.

\subsection{Stochastic sources in LI basis computation}
We apply the same approximation to the calculation of $M$ in equation \eqref{eq:G_M_transformer}, and obtain
\begin{equation}\label{eq:M_stoch_pos}
\begin{aligned}
M^{l}_{j,j'} \approx \frac{1}{R}\frac{1}{\vert \dataset \vert}\frac{1}{T} \sum_{x \in \dataset} \sum^R_{r=1} \sum^T_{t=1} \sum^{d^{l+1}}_{i=1} \frac{\partial {\left(\sum^T_{s=1} \phi_{r,s}(x) \hat{f}^{l+1}_{i,s}(\tilde{\mathbf{f}}^{l}(x))\right)}}{\partial \tilde{f}^{l}_{j,t}(x)}
\frac{\partial {\left(\sum^T_{s'=1} \phi_{r,s'}(x) \hat{f}^{l+1}_{i,s'}(\tilde{\mathbf{f}}^{l}(x))\right)}}{\partial \tilde{f}^{l}_{j',t}(x)}\,.\\
\end{aligned}
\end{equation}

In this case we could also apply the stochastic sources approach to the sum over index $i$, either
separately with two sets of sources, or by using the same sources for both sums. Testing the
accuracy of this option however we found that the error due to stochastic sources was large, even
when using $R_i=d^{l+1}$ sources for the $i$ index. Thus we use the full sum over $i$ and
stochastic sources only for the sum over $s$.

\subsection{Number of stochastic sources}
In practice we find that using $R=1$ stochastic source is often sufficient.
For MLP layers this is actually exact (because gradients between different positions are zero),
and for attention layers we find that the error from using $R=1$ is typically small compared to
the error introduced by the finite dataset size. We use $R=1$ in all experiments in this paper.

Most importantly, given a computational budget, $R$ trades off directly against the number of data
points $\vert\mathcal{D}\vert$, it is always better to use more data points than more sources. This
is because increasing $R$ samples the same datapoint with a new source, while increasing
$\vert\mathcal{D}\vert$ samples a new datapoint \textit{and} a new source.

We can also see this tradeoff in the stochastic sources error
estimator \citep{Dong_1994}
\begin{equation}\label{eq:stoch_var}
\begin{aligned}
\sigma^2_{\left(E^{l+1,l}_{i,j}\right)^2} &= \frac{2}{R}\frac{1}{\vert \dataset \vert}\sum^R_{r=1}\sum_{x \in \dataset} \frac{1}{T^2} \left(\sum^T_{s,s'=1}{\left(A^l_{(i,s),j}(x)\right)}^2{\left(A^l_{(i,s'),j}(x)\right)}^2-\sum^T_{s=1}{\left(A^l_{(i,s),j}(x)\right)}^4\right)
\end{aligned}
\end{equation}
which scales equally with $1/R$ and $1/\vert \dataset \vert$.

\section{Pseudocode for LIB and IG in transformers}
\label{appendix/pseudocode}
In this section, we show pseudocode for calculating the LIB basis (see Section \ref{subsec:LIB})
and averaged attributions (see Section \ref{subsec:IG}) in transformers. The pseudocode shows the
non-stochastic version of the code, i.e. without stochastic sources explained in appendix
\ref{appendix/stoch_sources}. We provide the full source code, with all functionality,
at \url{https://github.com/ApolloResearch/rib}.

In the rest of the paper, matrix vector products are defined left to right, as in $\hat{f}^l=C^l
f^l$, keeping with common mathematical convention. In this section, they are defined right to left
instead, as in $\hat{f}^l=f^l C^l $, keeping with the Pytorch convention used in the code base. As a
result, the ordering of matrices is often flipped compared to the definitions in the main text.
Similarly, we define $\hat{E}$ to be the RMS of attributions in the main text, but in the code base
(and the pseudocode) we define $E$ to be the sum of squares (without the square root).

\subsection{Local Interaction Basis in transformers}
Algorithm \ref{alg:lib} shows pseudocode for finding the Local Interaction Basis in a transformer. In the actual code base, we include the option to use stochastic sources in this algorithm to lower computing costs, see Appendix \ref{appendix/stoch_sources}.
\begin{algorithm}
\caption{Pseudocode for finding the LI basis in Transformers.}\label{alg:lib}
\begin{algorithmic}
\STATE \textbf{Output:} Returns interaction transformation matrix $C^l$ in layer $l$ for $l\in\{0,\ldots,l_{\text{final}}\}$ 
\FOR{layer $l\in\{0,\ldots,l_{\text{final}}\}$}
    \STATE ${G'}^l\gets \mathbf{0}^l_{(d^l+1,d^l+1)}$
    \STATE $H^l\gets \mathbf{1}^l_{(d^l+1,d^l+1)}$
    \FOR{$x$ in dataset $\dataset$, $t\in\{1,\ldots,T\}$, $j\in\{0,\ldots,d^l\}$}
        \STATE $H^l_{j,-1}\mathrel{-}=\frac{1}{\vert\dataset\vert T}\sum_{x,t}f^l_{x,t,j'}$
        \STATE $G^l_{j,j'} \mathrel{+}= \frac{1}{\vert\dataset\vert T} \text{einsum}(``xtj,xtj'\to jj'", f^l, f^l)$
    \ENDFOR
\ENDFOR
\FOR{layer $l\in\{l_{\text{final}},\ldots,0\}$}
    \STATE $G^l    \gets    H^l {G'}^l{(H^l)}^T$
    \STATE $D'^l$, $U'^l \gets \text{eigendecompose}(G^l[1:,1:])$
    \STATE $U^l \gets \mathbf{1}^l_{(d^l+1,d^l+1)}$
    \STATE $U^l[1:,1:] = U'^l$
    \IF {$l == l_{\text{final}}$}
        \STATE $C^l\gets H^l U^l$
    \ELSE
        \STATE $D^l=\text{remove\_tiny\_eigenvals\_and\_diagonalize}(D'^l)$
        \STATE $M'^l_{j,j'}\gets \mathbf{0}_{(d^l+1,d^l+1)}$
        \FOR{$x$ in dataset $\dataset$, $s, t\in\{1,\ldots,T\}$, $i\in\{0,\ldots,d^{l+1}\}$, $j\in\{0,\ldots,d^l\}$}
            \STATE $J^l_{x,i,s,j,t} \gets \frac{\partial  \big(\sum_{i'} C^{l+1}_{i,i'} f^{l+1}_{x,i',s}\big) }{\partial ( f^l_{x,j,t})}$
        \ENDFOR
        \STATE $M'^l_{j,j'} \mathrel{+}= \frac{1}{\vert \dataset\vert T} \text{einsum}(``xisjt,xisj't\to jj'", J^l, J^l)$
        \STATE $M^l\gets {D^l}^\frac{1}{2} {U^l}^T {H^l}^{-1} M'^l {{H^l}^{-1}}^T U^l {D^l}^\frac{1}{2}$
        \STATE $\Lambda'^l$, $V'^l \gets \text{eigendecompose}(M^l[1:,1:])$
        \STATE $\Lambda^l\gets \text{remove\_tiny\_eigenvals\_and\_diagonalize}(\Lambda'^l)$
        \STATE $V^l \gets \mathbf{1}^l_{d^l+1,d^l+1}$
        \STATE $V^l[1:,1:] = V'^l$
        \STATE $C^l\gets H^l {U^l} ({D^l}^\frac{1}{2})^+ V^l {(\Lambda^l)}^\frac{1}{2}$
    \ENDIF
\ENDFOR
\end{algorithmic}
\end{algorithm}

\subsection{Edges in transformers}
Algorithm \ref{alg:lib} shows pseudocode to calculate the edges quantifying interaction strength between features in the Local Interaction basis in a transformer. In the actual code base, we include the option to use stochastic sources in this algorithm to lower computing costs, see Appendix \ref{appendix/stoch_sources}.
\begin{algorithm}
\caption{Pseudocode for computing integrated gradient attributions in transformers.}\label{alg:edge}
\begin{algorithmic}
\STATE \textbf{Output:} Returns interaction edges $\hat{E}^l$ in layers $l\in\{0,\ldots,l_{\text{final}}\}$ 
\FOR{layer $l\in\{0,\ldots,l_{\text{final}}-1\}$}
    \STATE $E^{l+1,l}_{i,j}\gets \mathbf{0}_{(d^{l+1}+1,d^l+1)}$
    \FOR{$x$ in dataset $\dataset$, $s\in\{1,\ldots,T\}$, $i\in\{0,\ldots,d^{l+1}\}$, $j\in\{0,\ldots,d^l\}$}
        \STATE $A^{l+1,l}_{x,i,s,j}\gets \mathbf{0}_{(\vert\dataset\vert,d^{l+1}+1,T, d^l+1)}$
            \FOR{$\alpha \in \{0,\ldots,1\} \text{\quad(step size $p$)}$}
            \STATE $p'=0.5p$ if $\alpha=0$ or $\alpha=1$ else $p'=p$
            \STATE $\hat{f}^{l+1}(\alpha \hat{f}^{l})_{x,i,s}\gets  {\left(f^{l+1}(\alpha \hat{f}^{l}(C^l)^+)C^{l+1} \right)}_{x,i,s} $
            \FOR{$t\in \{ 1,\ldots,T \} $}
                \STATE $A^{l+1,l}_{x,i,s,j} \mathrel{+}= \text{einsum}(``xisjt,xjt\to xisj", p'\frac{\partial \left(\sum_b\hat{f}^{l+1}(\alpha \hat{f}^{l})_{x,i,s}\right)}{\partial \alpha \hat{f}^{l}_{x,j,t}}, \hat{f}^l(X)_{x,j,t})$
            \ENDFOR
        \ENDFOR
    \STATE $\hat{E}^l_{i,j} \mathrel{+}= \frac{1}{\vert \dataset\vert T}\sum_{x,s}\left(A^{l+1,l}_{x,i,s,j}\right)^2$
    \ENDFOR
\ENDFOR
\end{algorithmic}
\end{algorithm}

\section{Alternative: Global Interaction Basis}
\label{appendix/gradient_flow}
In addition to the Local Interaction Basis (LIB) described in Section \ref{subsec:LIB}, we also
experimented with what we call the Global Interaction Basis (GIB). This is an earlier variant of LIB
proposed in \citet{apolloTheory}.
To compute the GI basis we adjust
second transformation to align the basis of activations in layer $l$ with the basis
in the final layer $l_{\text{final}}$ instead of layer $l+1$. In practice we found similar
results for the LIB and GIB, and thus focus on the LIB in the main text.

The difference is only in equation \eqref{eq:M} which, for GIB, is
\begin{equation}\label{eq:M_gradient_flow}
{M}^{l}_{jj'} \coloneq
    \frac{1}{\vert \dataset \vert}
    \sum_{x \in \dataset}\sum_{i=0}^{d^{l_\text{final}}}
        \frac
            {\partial \hat{f}^{{l_\text{final}}}_i({\tilde{\mathbf{f}}}^{l}(x))}
            {\partial {\tilde{f}}^{l}_j(x)}
        \frac
            {\partial  \hat{f}^{{l_\text{final}}}_i({\tilde{\mathbf{f}}}^{l}(x))}
            {\partial {\tilde{f}}^{l}_{j'}(x)}\,.
\end{equation}

\subsection{GIB results}
We evaluated GIB on the modular addition and Tinystories-1M transformers. We found that the results
were similar to LIB, resulting in similar sparsity and interaction graphs.

\begin{figure}[ht]
    \centering
    \includegraphics[width=0.49\textwidth,trim=0 0 0 0.02\textwidth,clip]{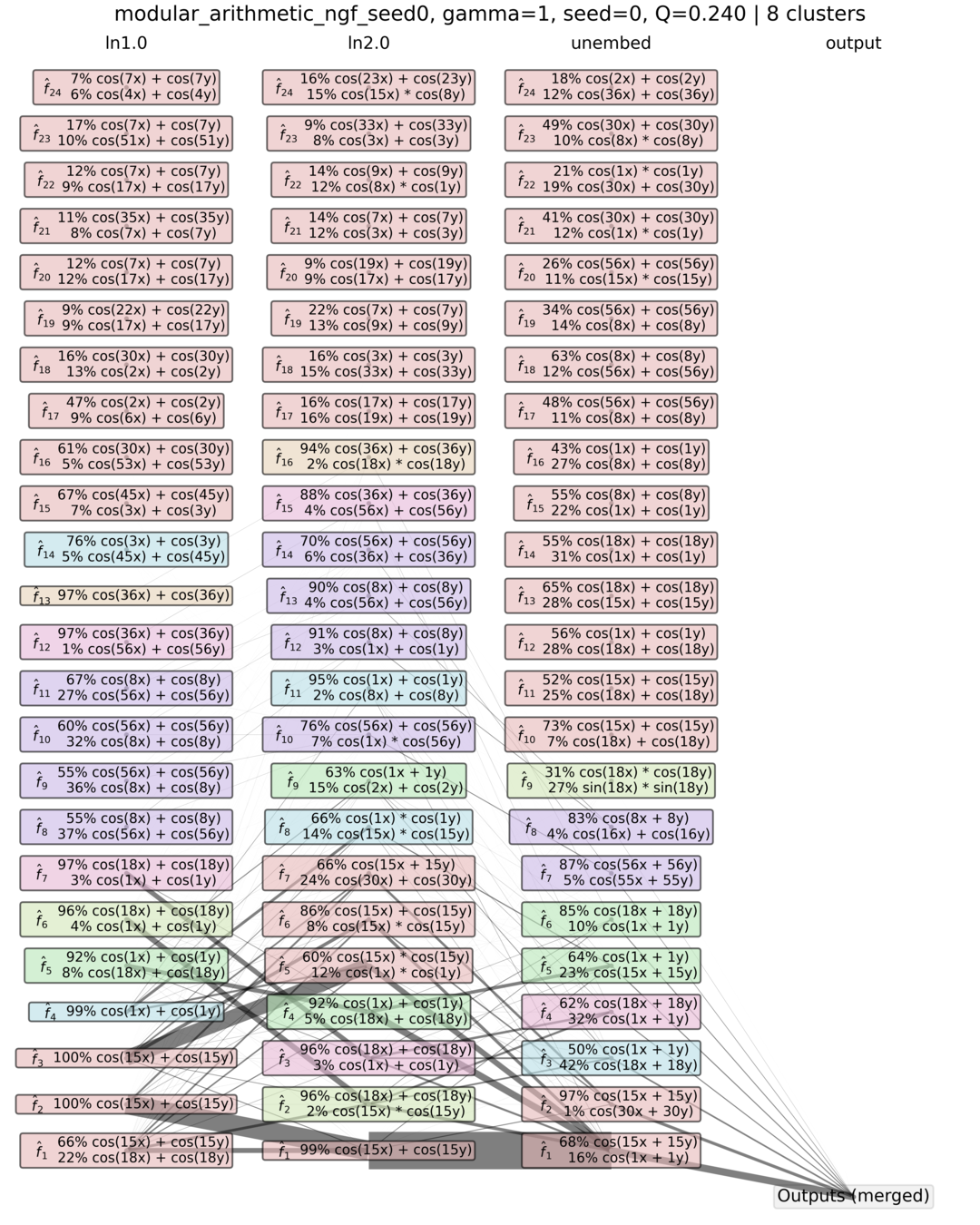}
    \includegraphics[width=0.49\textwidth,trim=0 0 0 0.02\textwidth,clip]{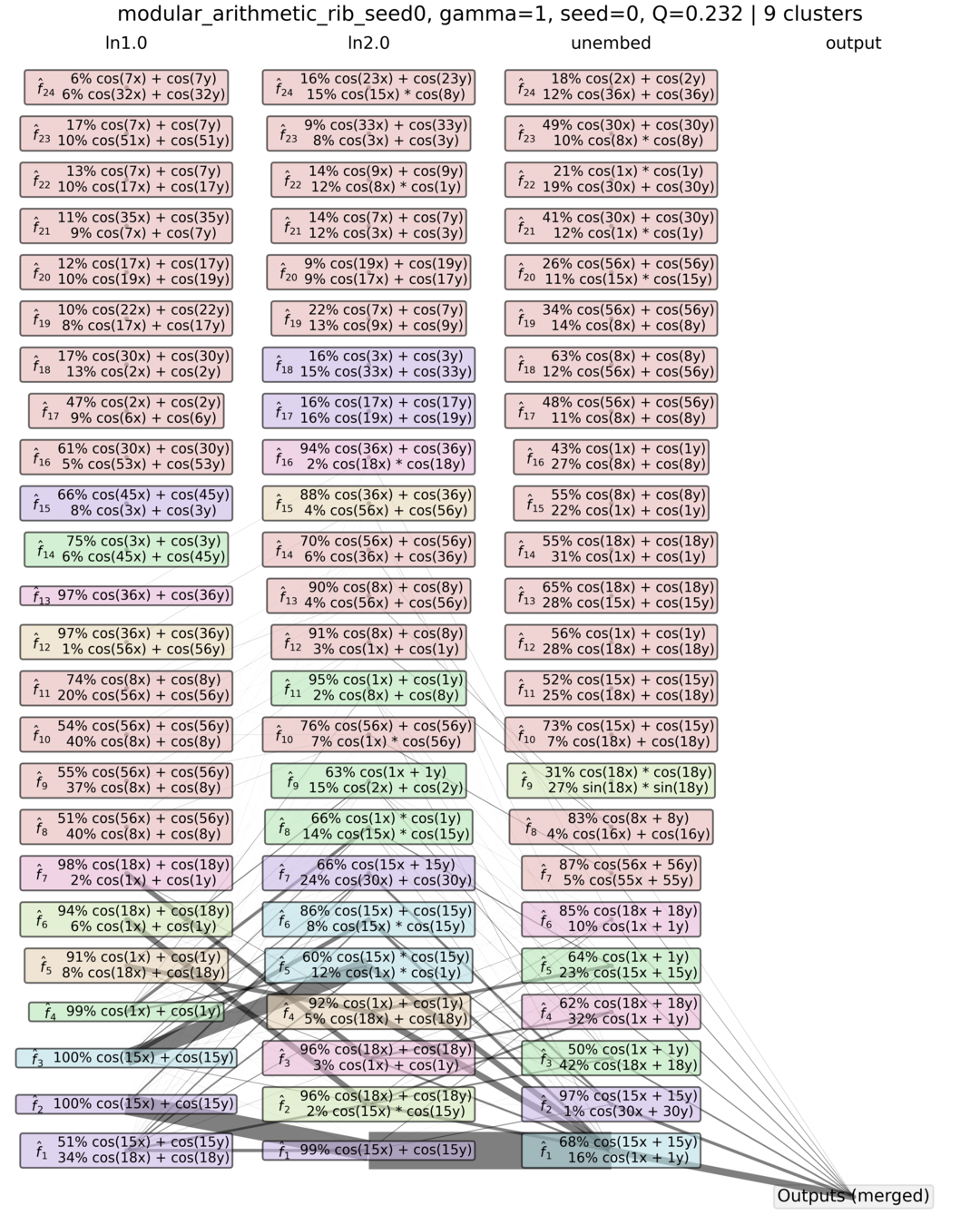}
    \caption{GIB (left) and LIB (right) interaction graph for the modular addition transformer (seed-0).}
    \label{fig:modular_arithmetic_lib_vs_gib}
\end{figure}

As an example we show the interaction graph for the modular addition transformer (seed-0) in Figure
\ref{fig:modular_arithmetic_lib_vs_gib} for GIB and LIB. While the nodes are not exactly identical, the
features are very similar and no graph is clearly more interpretable.

Performing the edge-ablation test for both bases we find that the GIB requires 131 and 82 edges
across the attention and MLP layers, respectively. This is essentially identical to the LIB results
(133 and 82 edges, respectively).

\section{Fourier decomposition of activations in modular addition}
\label{appendix/modular_addition_math}
In this section we briefly explain the Fourier decomposition we use to interpret the activations
in the modular addition transformer. This is not a novel idea and has been commonly employed in
the literature though we slightly improve on the representation used in \citet{nanda2023progress}.

Fundamentally we can obtain the activations for all $113^2$ possible inputs and plot the
any activation as a function of $x$ and $y$, $f(x,y)$. However, the activations tend to be
periodic in $x$ and $y$ due to the nature of this task, and it is more useful to
apply a discrete Fourier transform (Fast Fourier Transform, FFT). The set of
Fourier modes $f'_{k_x,k_y}$ contains the same information as the original set of activations
\begin{gather}
    f(x,y) = \sum_{k_x=-56}^{56} \sum_{k_y=-56}^{56} f'_{k_x,k_y} e^{2\pi i \frac{k_x x + k_y y}{113}}\,.
    \label{eq:fourier_decomposition}
\end{gather}

The Fourier modes $f'_{k_x,k_y}$ tend to be sparse, i.e. $f'$ is zero for most values of $k_x$
and $k_y$. Thus we can describe (label) activations as a small list of frequencies. The rest of
this section discusses how to transform the set of complex numbers $f'_{k_x,k_y}$ into a list of
sinusoidal terms.

First we separate the amplitude and phase information in $f'_{k_x,k_y}$. We write
\begin{gather}
    f'_{k_x,k_y} e^{2\pi i \frac{k_x x + k_y y}{113}} = a_{k_x,k_y} e^{2\pi i\frac{\phi_{k_x,k_y}}{113} + 2\pi i \frac{k_x x + k_y y}{113}}
    \label{eq:fourier_decomposition_exp}
\end{gather}
with real numbers $a_{k_x,k_y}$ and $\phi_{k_x,k_y}$. Next we make use of the fact that that every
non-zero frequency pair $(k_x,k_y)$ has four sign variations in equation \eqref{eq:fourier_decomposition}.
Because the input activations $f(x,y)$ were real, we know that $f'_{k_x,k_y} = f'^\ast(-k_x,-k_y)$,
and thus $a_{k_x,k_y} = a(-k_x,-k_y)$ and $\phi_{k_x,k_y} = -\phi(-k_x,-k_y)$. We can make use of
these relations to turn equation \eqref{eq:fourier_decomposition_exp} into a series of sinusoidal terms
\begin{equation}
    \begin{aligned}
    f(x,y) &= a_{0, 0} + \sum_{k_x=1}^{56} 2a_{k_x,0} \cos\left({\scriptstyle 2\pi \frac{k_x x  + \phi_{k_x,0}}{113}}\right)
                      + \sum_{k_y=1}^{56} 2a_{0,k_y} \cos\left({\scriptstyle 2\pi \frac{k_y y  + \phi_{0,k_y}}{113}}\right)
    \\&+\sum_{k_x=1}^{56} \sum_{k_y=1}^{56} 2a_{k_x,k_y} \cos\left({\scriptstyle 2\pi \frac{k_x x + k_y y + \phi_{k_x,k_y}  }{113}}\right)
                                         + 2a_{k_x,-k_y} \cos\left({\scriptstyle 2\pi \frac{k_x x - k_y y + \phi_{k_x, -k_y}}{113}}\right)
                                         \,.\label{eq:fourier_decomposition_cos}
    \end{aligned}
\end{equation}
Because all these terms are orthogonal we can compute the variance of $f(x,y)$ that is explained
by each term directly from their coefficients. As a shorthand notation we write $\cos(k_x x + k_y y)$
to denote $\cos\left({\scriptstyle 2\pi \frac{k_x x + k_y y}{113}}\right)$.
We use this to label the features in the modular
addition analysis, Section \ref{subsec:modular_addition} and appendix
\ref{appendix/modular_addition_all_seeds}.

The ${\scriptstyle k_x + k_y}$ and ${\scriptstyle k_x - k_y}$ terms usually give a simple
description of the activations, with one of the terms being zero. But sometimes we notice the
respective coefficients have nearly the same value, with the same or with opposite signs. We
notice that this corresponds to the trigonometric identities
\begin{gather}
    \frac{1}{2} \left(\cos(x+y) + \cos(x-y)\right) = \cos(x) \cos(y)
    \\
    \frac{1}{2} \left(\cos(x+y) - \cos(x-y)\right) = -\sin(x) \sin(y)
\end{gather}
and thus dynamically switch from the ${ \cos(k_x + k_y)}$ and ${ \cos(k_x - k_y)}$
notation to the ${ \cos(k_x) \cos(k_y)}$ and ${ \sin(k_x) \sin(k_y)}$
notation when the latter provides a more sparse (more compressed) description.

\section{More modular addition results}
\label{appendix/modular_addition_all_seeds}
\begin{figure}
    \centering
    \adjincludegraphics[width=0.49\textwidth,trim={0 0 0 0.05\height},clip]{modular_arithmetic_rib_seed0-delta0.001-gamma1-sortingrib-graph.png}
    \adjincludegraphics[width=0.49\textwidth,trim={0 0 0 0.05\height},clip]{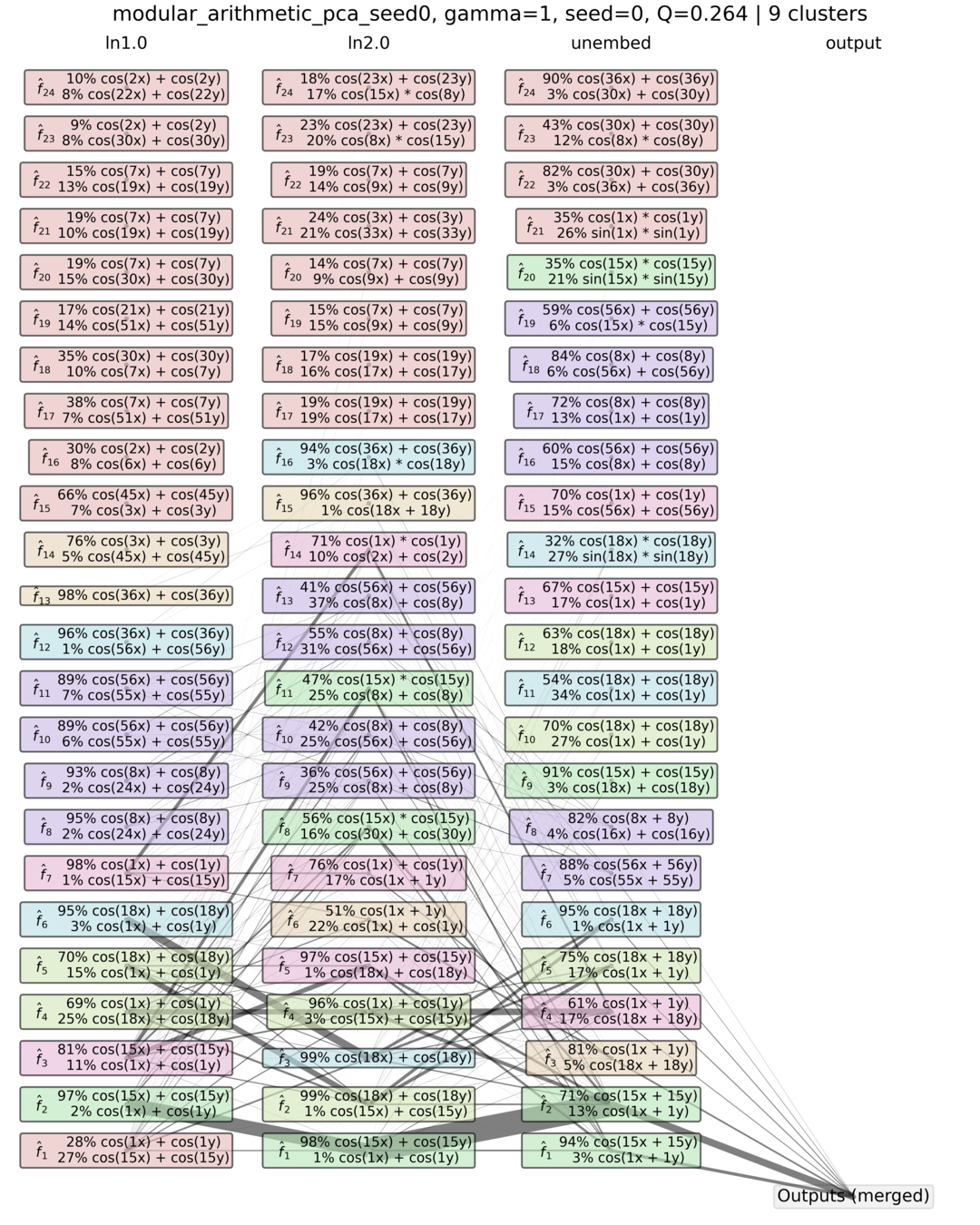}
    \caption{RIB (left) and PCA (right) interaction graphs of a modular addition transformer (seed-0).}
    \label{fig:modular_arithmetic_rib_seed0}
\end{figure}

\begin{figure}
    \centering
    \adjincludegraphics[width=0.49\textwidth,trim={0 0 0 0.28\height},clip]{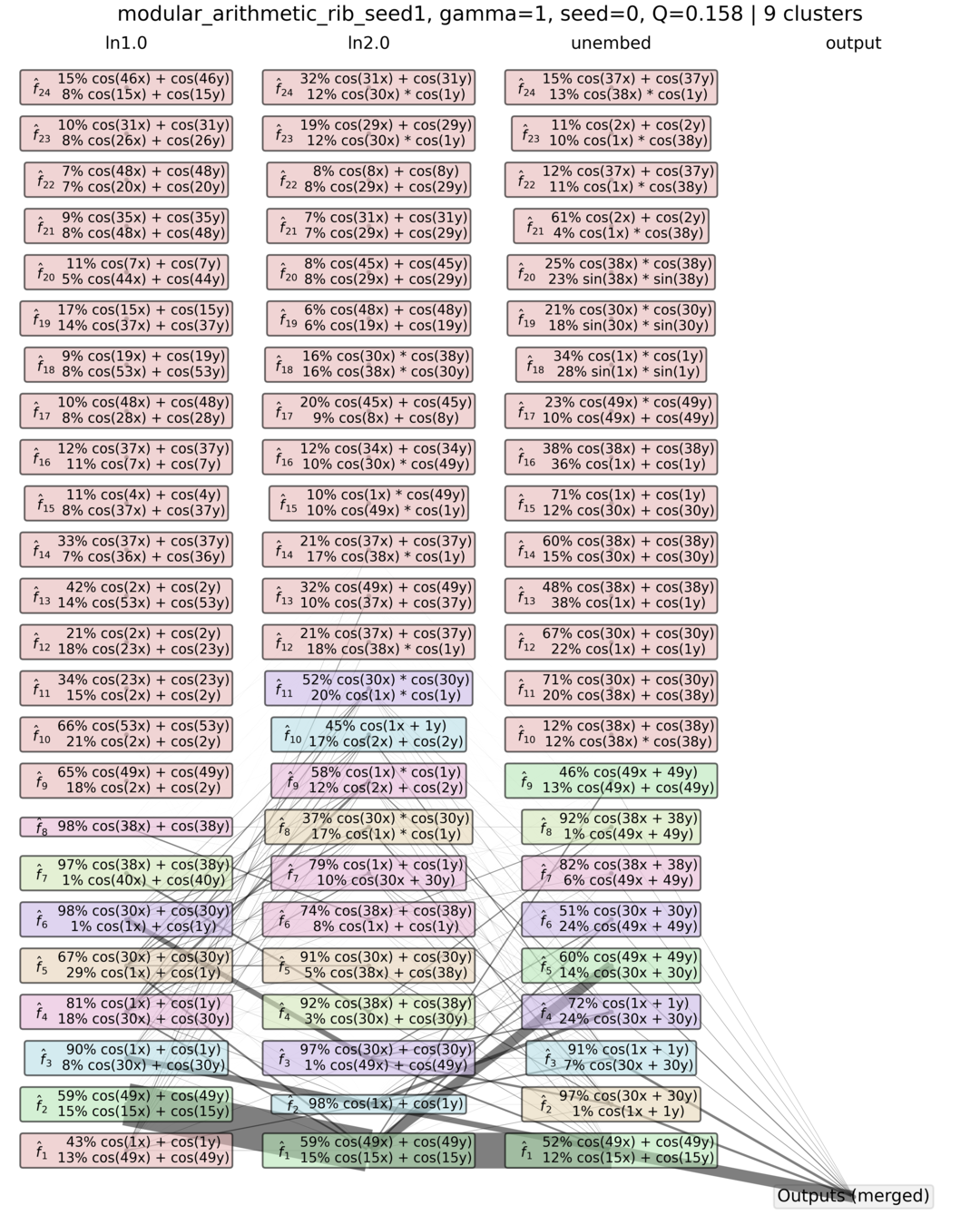}
    \adjincludegraphics[width=0.49\textwidth,trim={0 0 0 0.28\height},clip]{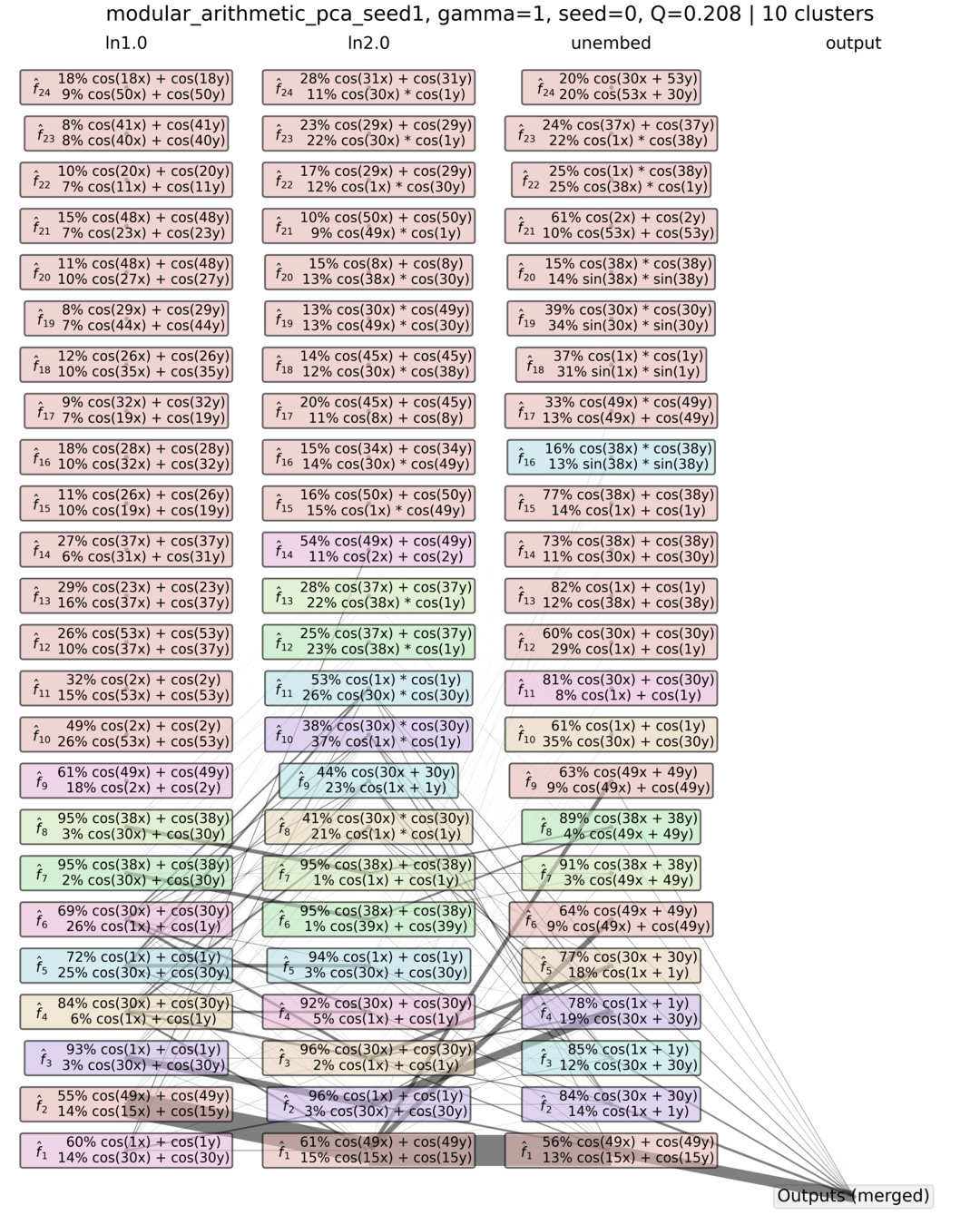}
    \caption{RIB (left) and PCA (right) interaction graphs of a modular addition transformer (seed-1).}
    \label{fig:modular_arithmetic_rib_seed1}
\end{figure}

\begin{figure}
    \centering
    \adjincludegraphics[width=0.49\textwidth,trim={0 0 0 0.28\height},clip]{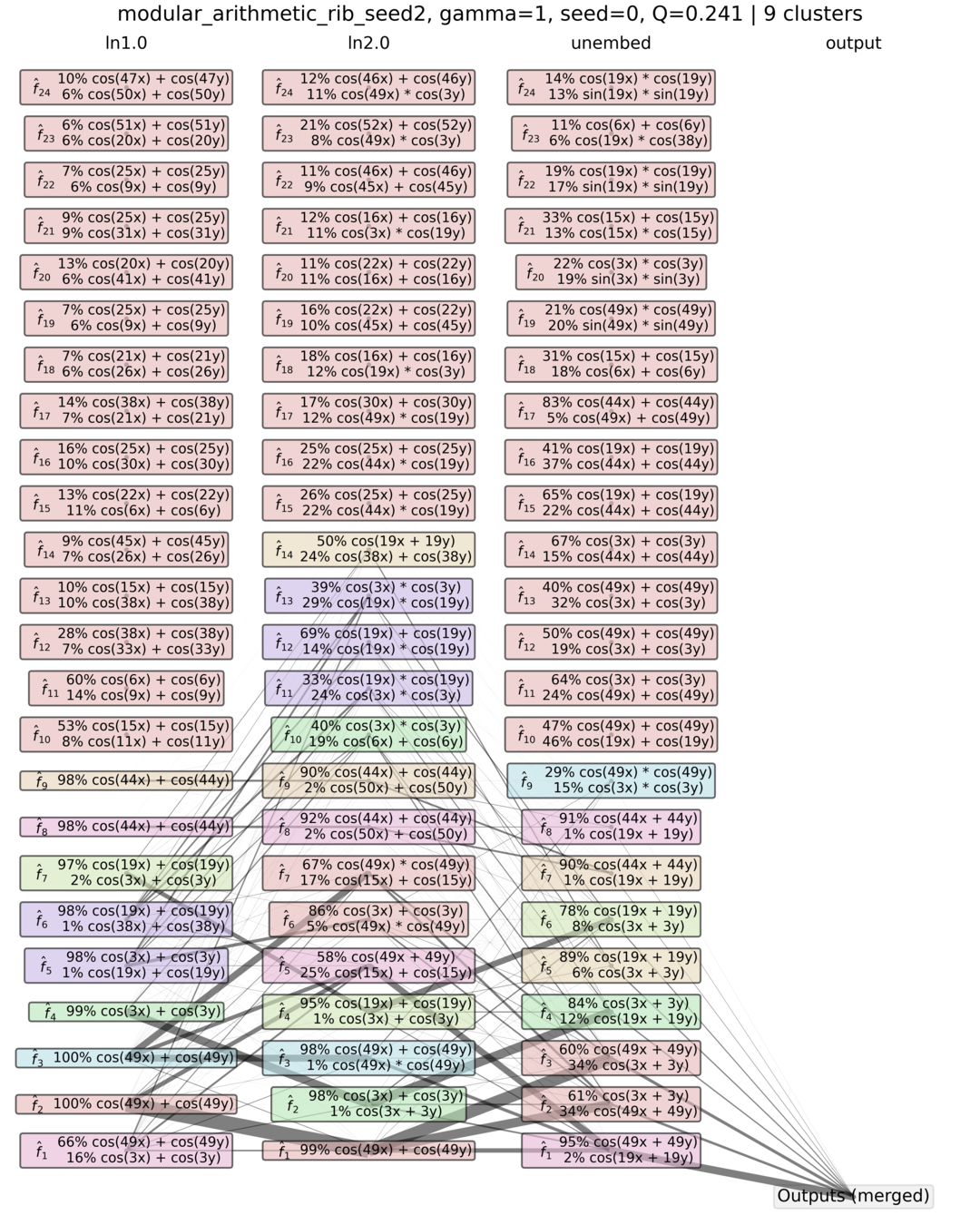}
    \adjincludegraphics[width=0.49\textwidth,trim={0 0 0 0.28\height},clip]{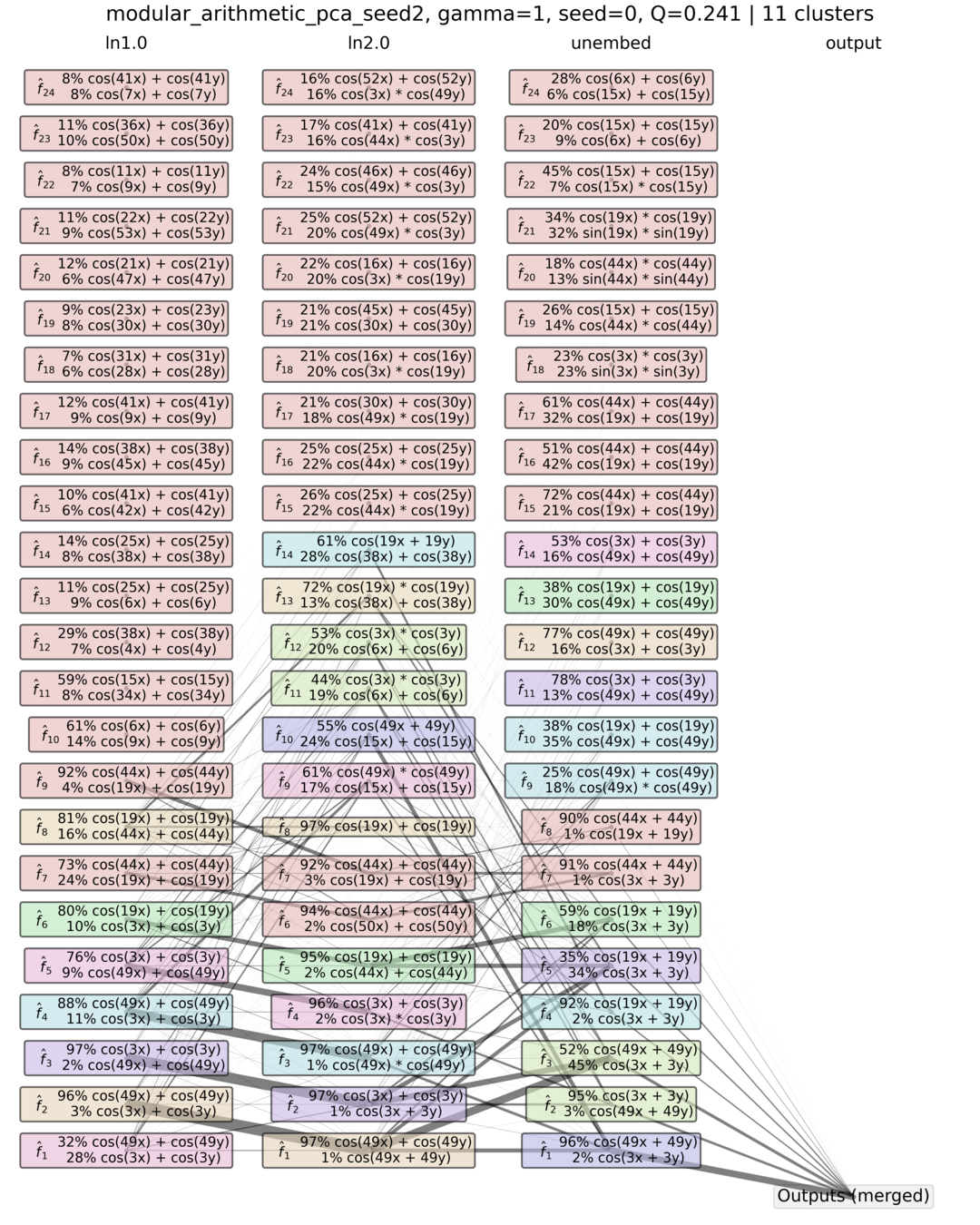}
    \caption{RIB (left) and PCA (right) interaction graphs of a modular addition transformer (seed-2).}
    \label{fig:modular_arithmetic_rib_seed2}
\end{figure}

\begin{figure}
    \centering
    \adjincludegraphics[width=0.49\textwidth,trim={0 0 0 0.05\height},clip]{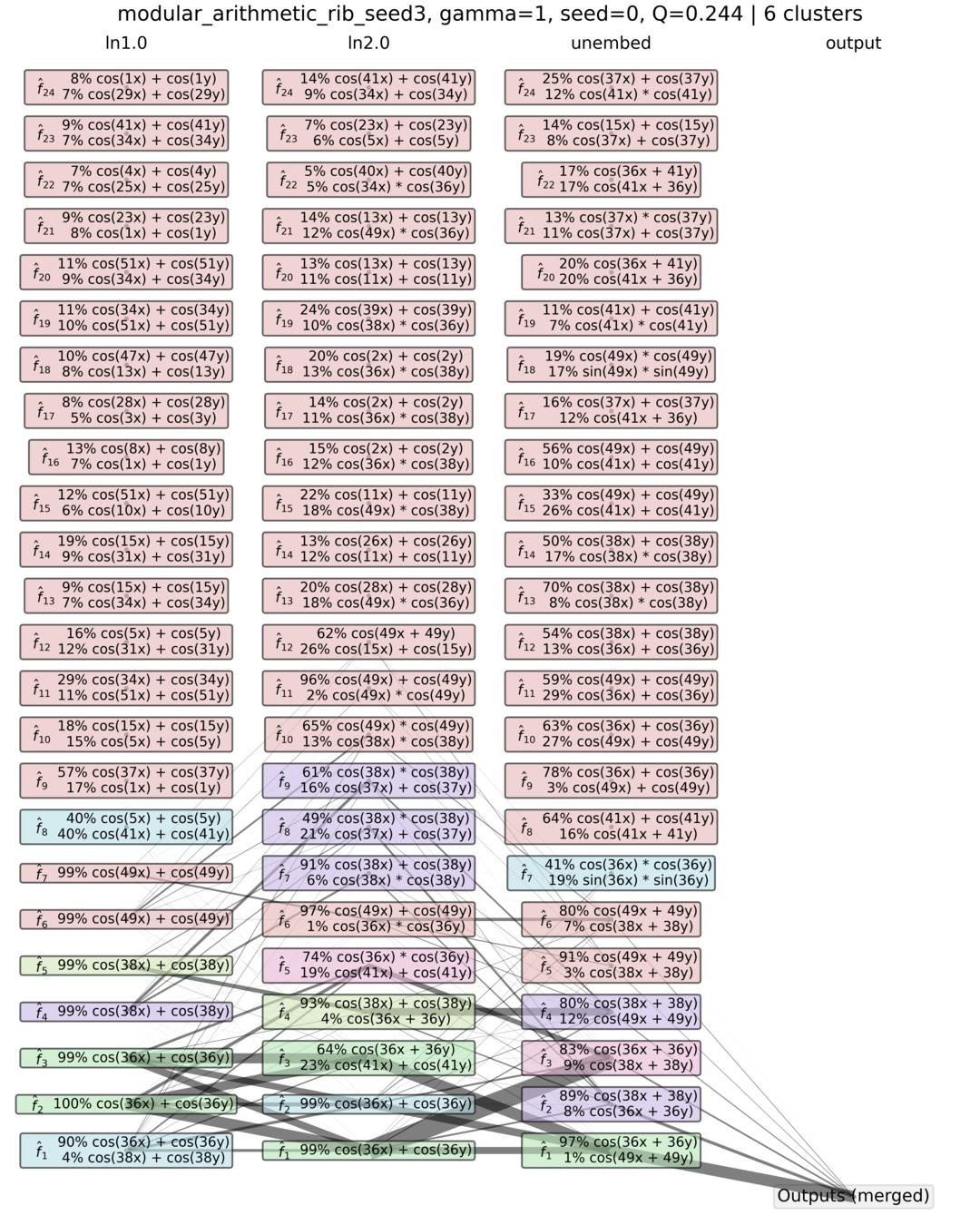}
    \adjincludegraphics[width=0.49\textwidth,trim={0 0 0 0.05\height},clip]{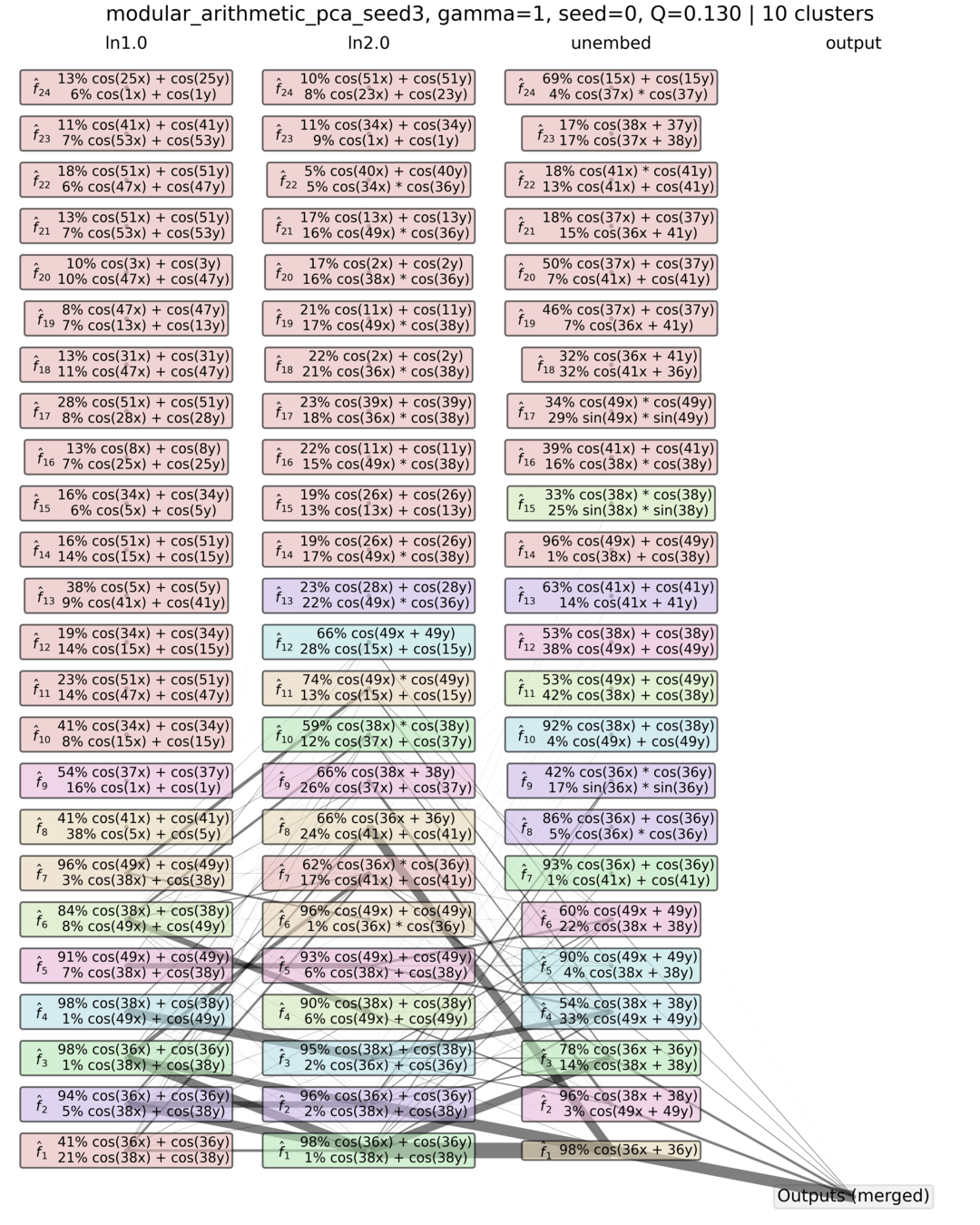}
    \caption{RIB (left) and PCA (right) interaction graphs of a modular addition transformer (seed-3).}
    \label{fig:modular_arithmetic_rib_seed3}
\end{figure}

\begin{figure}
    \centering
    \adjincludegraphics[width=0.49\textwidth,trim={0 0 0 0.20\height},clip]{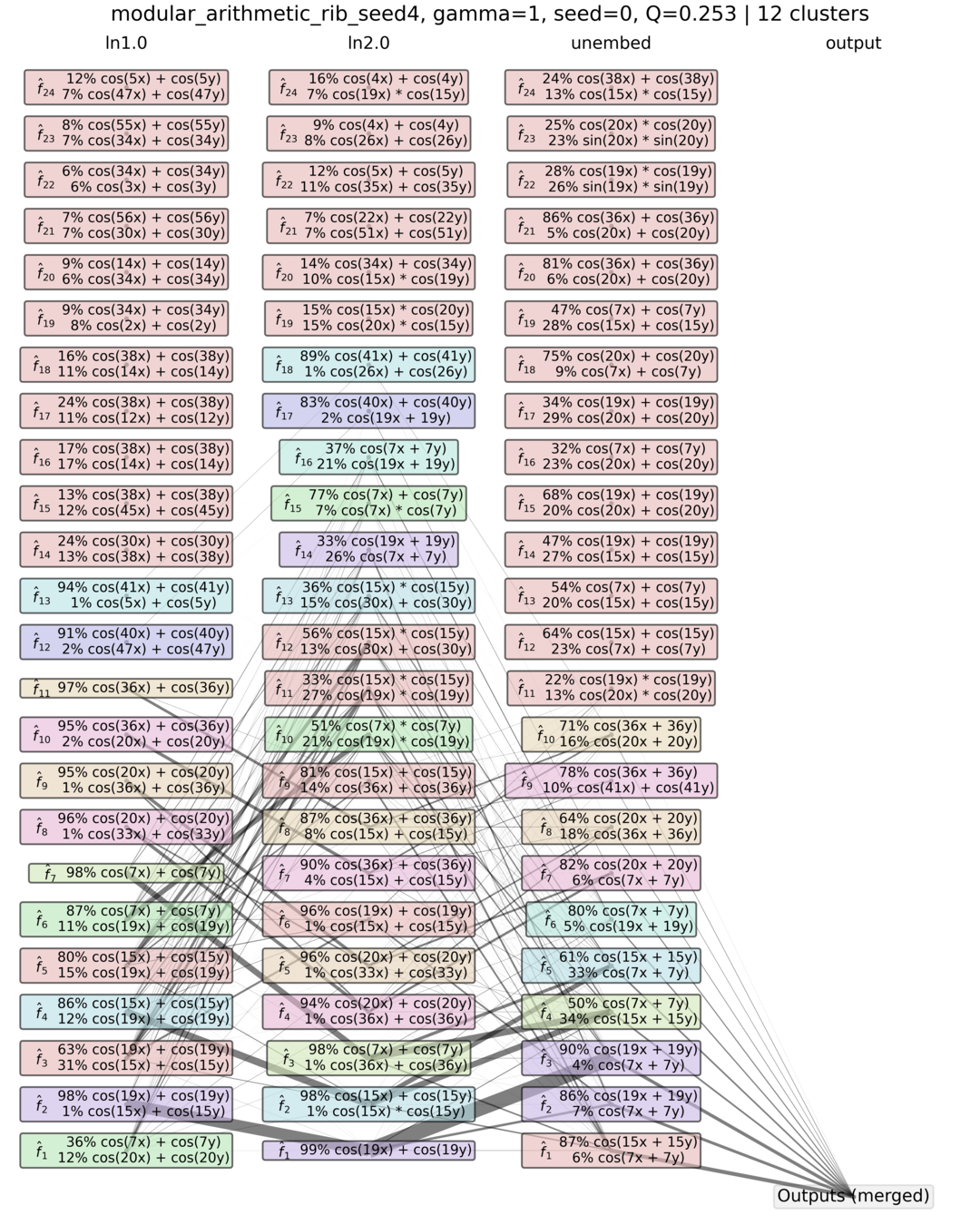}
    \adjincludegraphics[width=0.49\textwidth,trim={0 0 0 0.20\height},clip]{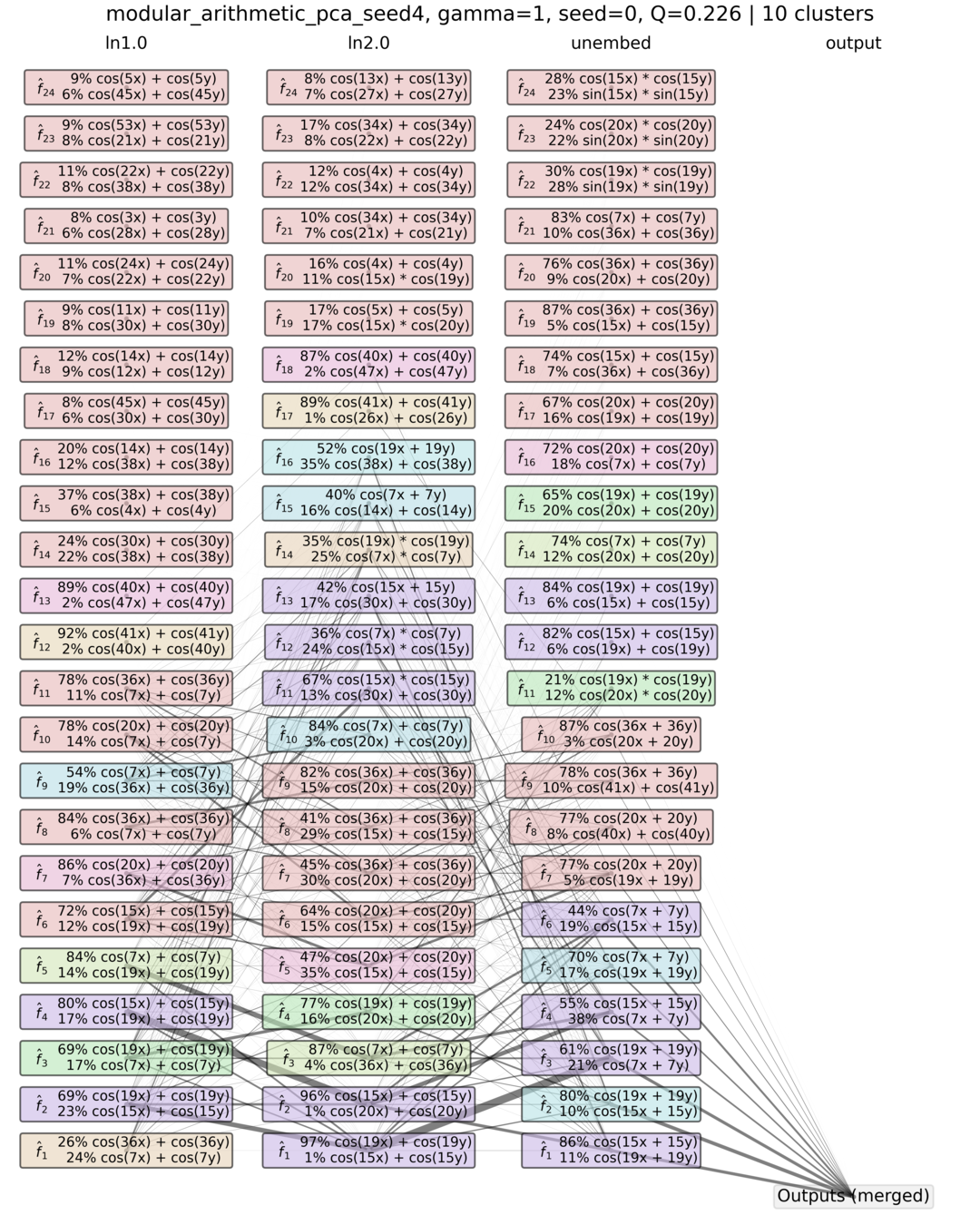}
    \caption{RIB (left) and PCA (right) interaction graphs of a modular addition transformer (seed-4).}
    \label{fig:modular_arithmetic_rib_seed4}
\end{figure}

Here we show interaction graphs for all LIB and PCA runs on all 5 versions of the modular addition
transformer. We show the RIB and PCA interaction graphs for seeds 0-4 in Figures
\ref{fig:modular_arithmetic_rib_seed0} to \ref{fig:modular_arithmetic_rib_seed4}.
Each plot shows four layers (before attention, between attention and MLP, after MLP, and the output)
and the nodes are sorted by the importance assigned by LIB or PCA, respectively. We label the
nodes with feature index $j$ as $\hat{f}_j$ (for practical reasons
we use $\hat{f}$ labels for both LIB and PCA, but
$\hat{f}$ in the PCA plots corresponds to $\tilde{f}$ in paper notation).

In these plots we show all features, including features which can be ablated
while maintaining $>99.9\%$ accuracy (coloured in grey) that we exclude
in the main text. We color the features by cluster, but do not
change the sorting based on the clustering to allow for easier comparison.
Additionally, we show the output layer and edges to it. Instead of showing all
113 output directions we combine them into a single node because the interactions
to all output directions are roughly identical.

In these comparisons we want to highlight the observation mentioned in section
\ref{subsubsec:mod_add_functional_relevance},
that some PCA features seem computationally-irrelevant---they do not connect to
future layers. We expect this to happen whenever a direction is computationally-irrelevant
but explains a significant amount of variance in the activations. LIB ignores such directions in
activation space. We find this phenomenon in two categories of layers:
\begin{enumerate}
    \item In feature layers just before a linear projection. For example the last layer before the
    unembedding (shown as the third column in the modular addition interaction graphs), or when we consider
    a layer right after the ReLUs (before the $W_{\text{out}}$ projection). The failure here
    is in some sense trivial: yes, PCA doesn't know about the linear projection, but one could
    argue that one should run PCA after linear projections.
    \item In layers that are \textit{not} immediately before a linear projection. For example, the
    residual stream between the attention and MLP layers (second column in the interaction graphs).
    Our model here is that the residual stream has some leftover information that is not used
    in the following layers. This is a more interesting case as there is no simple fix for PCA.
\end{enumerate}

We observe case 1 in all seeds of the modular addition transformer. Figures
\ref{fig:modular_arithmetic_rib_seed0} to \ref{fig:modular_arithmetic_rib_seed4} all
show features that are not connected to the outputs in the PCA basis.

Case 2 is less pronounced. We perform a manual inspection of the the first 25 features in the
second layer (between attention and MLP) for all seeds and bases. We go through Figures
\ref{fig:modular_arithmetic_rib_seed0} to \ref{fig:modular_arithmetic_rib_seed4} and
count the number of features have incoming edges (as a proxy for assigned relevance)
but no outgoing edges (as an indicator for functional relevance). This is a rough measure
and we expect some false positives but we can compare the two bases.
We indeed find more apparently-irrelevant features in the PCA basis (14) than in the LI basis (3).
Concretely we find PCA features 17 to 20 in seed-0, 15 and 16 in seed-1, 17 and 24 in seed-2, 22 in seed-3,
and 19 to 24 in seed-4 to be appear irrelevant. At the same time we find LIB features 18 in seed-0, feature
17 in seed-2, and feature 21 in seed-4 to be appear irrelevant.

\section{More LLM feature vizualizations}
\label{appendix/llm_features}
Here we show three more LIB features, randomly selected from block 6 of the GPT2-small model.
\begin{enumerate}
    \item Feature 82 does not appear to have a clear interpretation. Some dataset examples look similar, but there is no interpretation that explains a majority.
    \begin{center}
        \centering
    \includegraphics[width=0.45\textwidth]{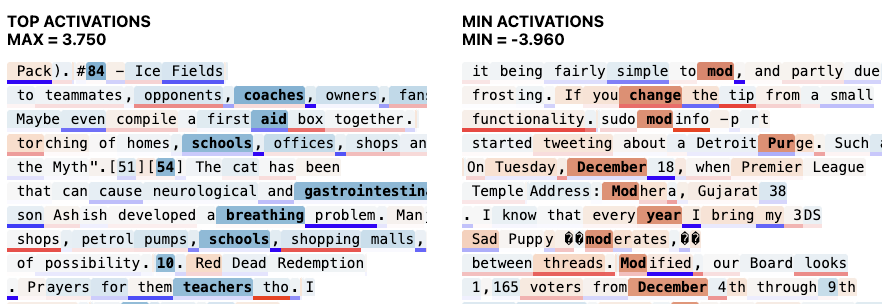}
    \includegraphics[width=0.45\textwidth]{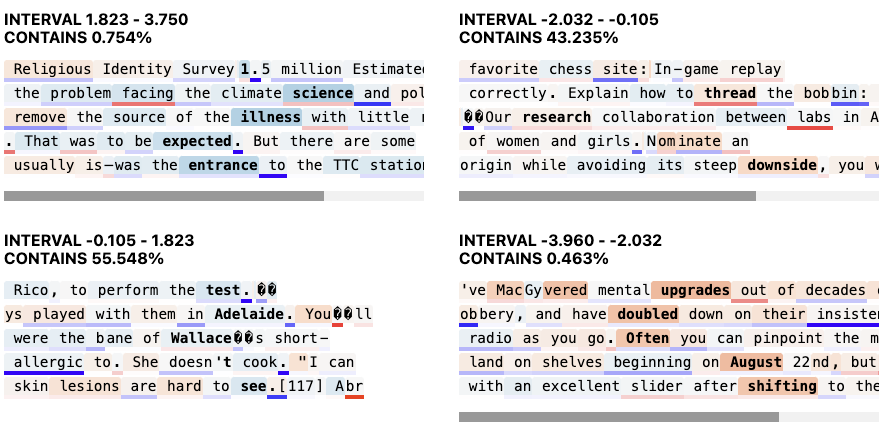}
    \end{center}

    \item Feature 304 seems to most strongly negatively activate on tokens related to watching, but this explanation does not generalize to intermediate or maximally positive activations.
    \begin{center}
        \centering
    \includegraphics[width=0.45\textwidth]{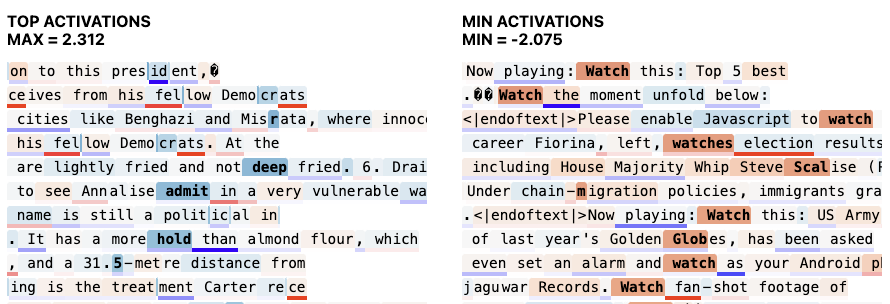}
    \includegraphics[width=0.45\textwidth]{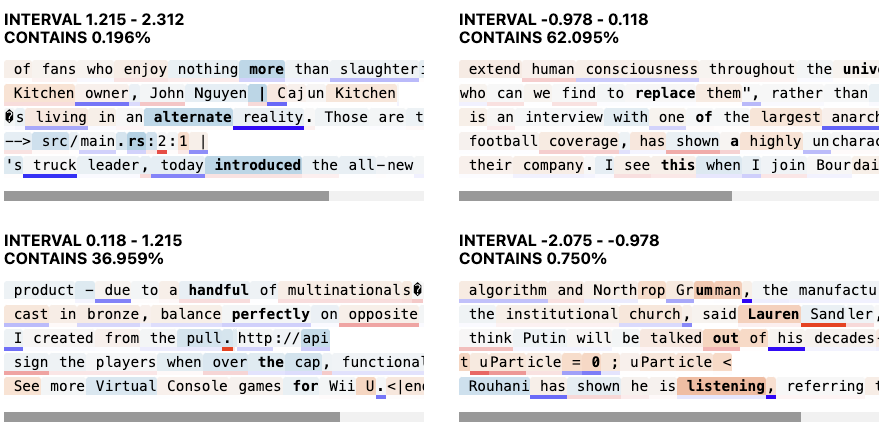}
    \end{center}
    \item Feature 477 does not appear to have a clear interpretation.
    \begin{center}
        \centering
        \includegraphics[width=0.45\textwidth]{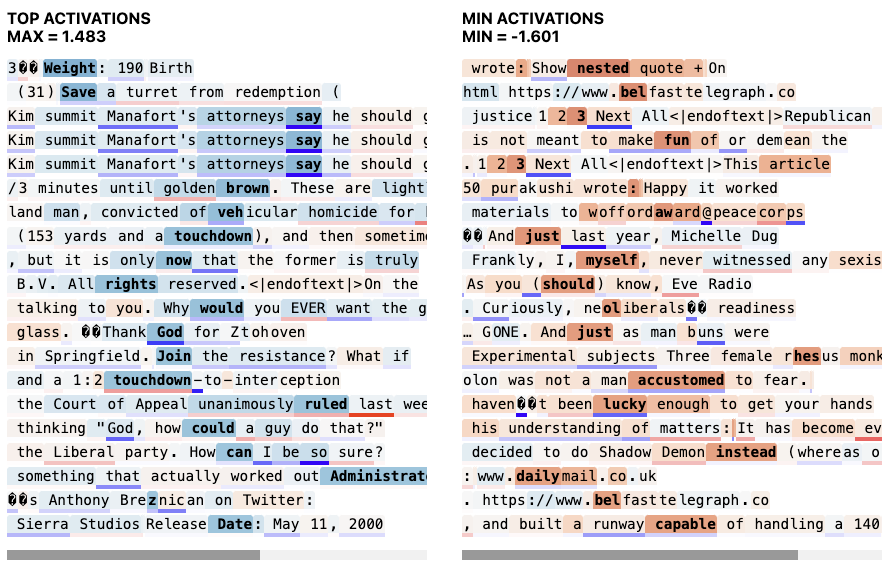}
        \includegraphics[width=0.45\textwidth]{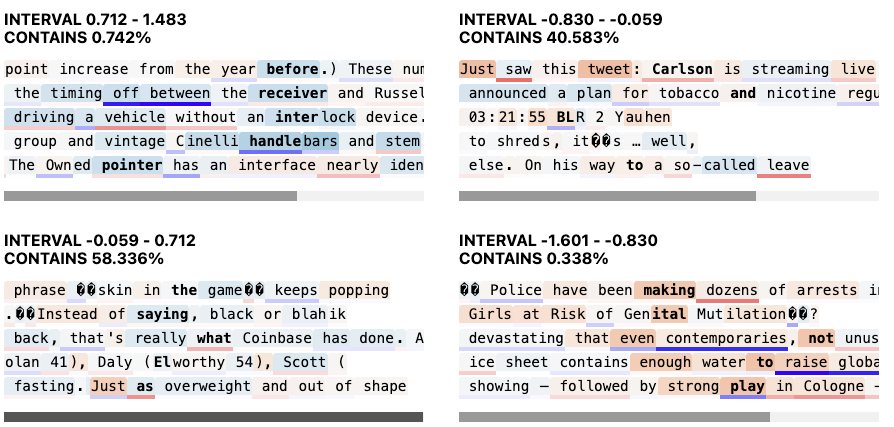}
    \end{center}
\end{enumerate}

\section{More LLM Ablation Results}\label{appendix/ablation-graphs}
\begin{figure}[h!]
    \centering
    \begin{subfigure}[b]{0.49\textwidth}
         \centering
         \includegraphics[width=\textwidth]{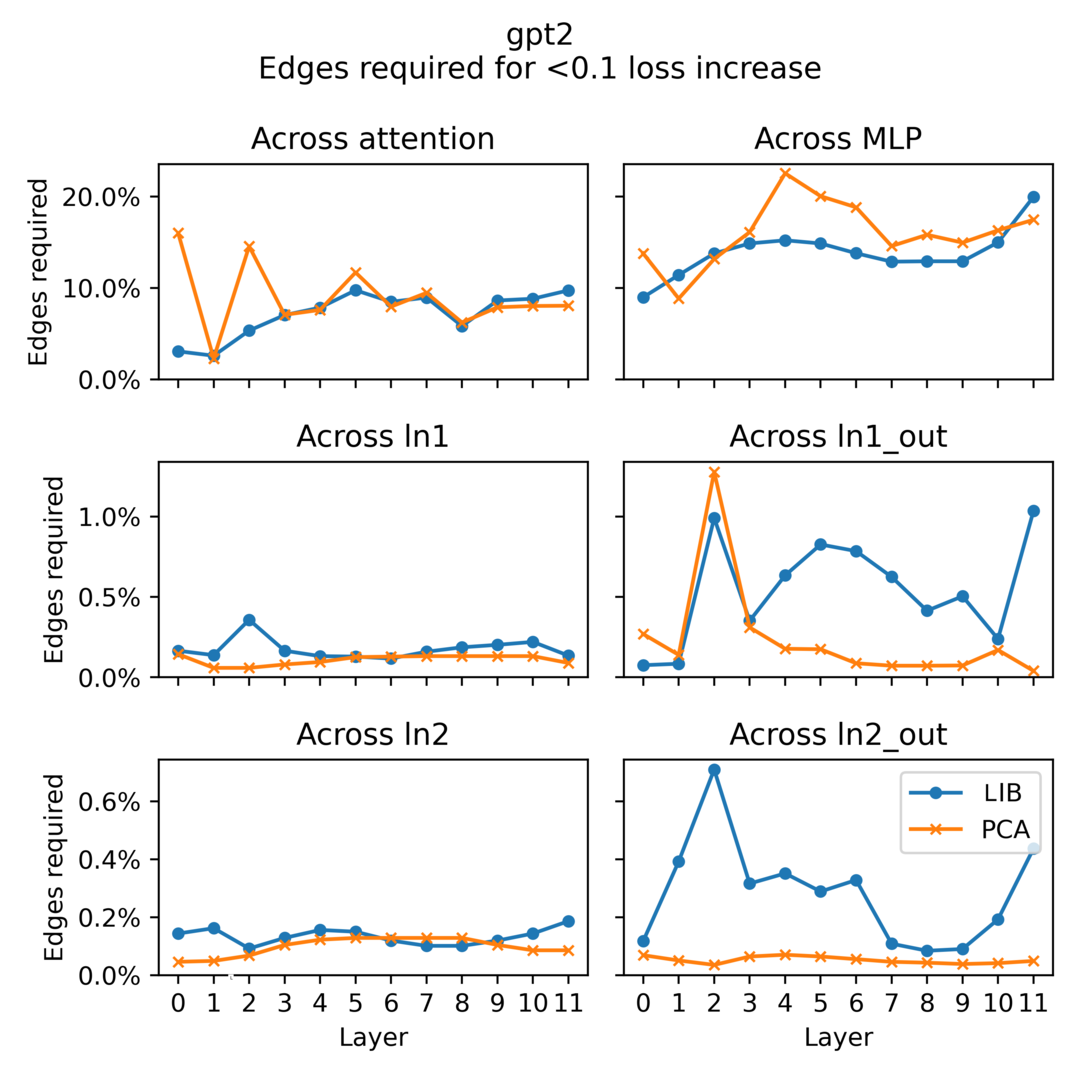}
         \caption{GPT2-small}
         \label{fig:gpt2-edge-abl-all}
    \end{subfigure}
    \begin{subfigure}[b]{0.49\textwidth}
         \centering
         \includegraphics[width=\textwidth]{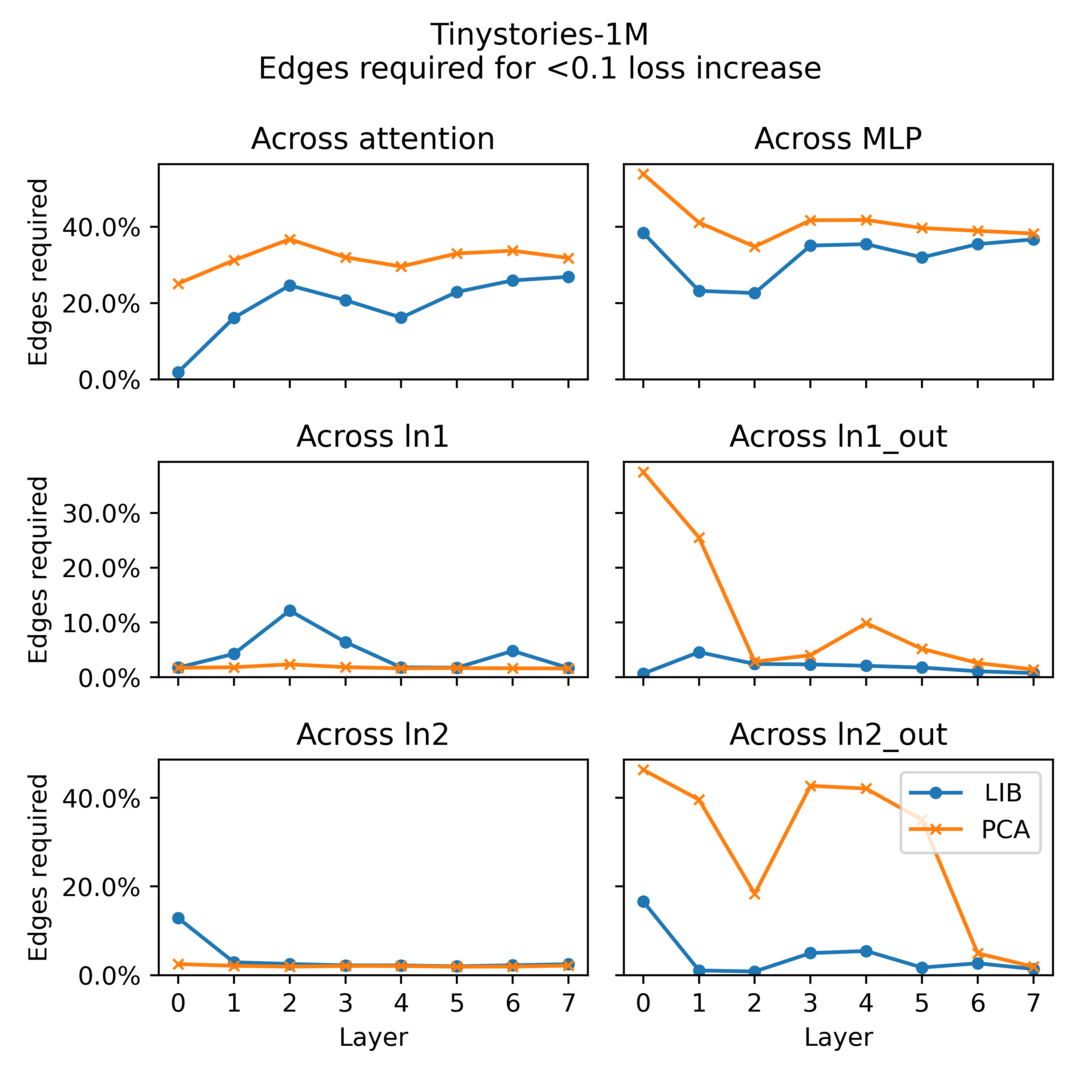}
         \caption{Tinystories-1M}
         \label{fig:tinystories-edge-abl-all}
    \end{subfigure}
    \caption{Edge ablation results on GPT2-small (left) and Tinystories-1M (right), for all layer types. Lower number of edges required is better, as it implies the graph is sparser.}
    \label{fig:appx-all-edge-ablations}
\end{figure}

Here we present edge ablation results for all layers in the GPT2-small and Tinystories-1M models.
Figure \ref{fig:appx-all-edge-ablations} shows the percentage of edges required (not ablated)
to maintain a loss increase below 0.1 for all layers in the models.

In the attention and MLP layers we find the LIB to often beat PCA for GPT2-small (as
shown in the main text), and LIB is sparser than PCA in all layers for Tinystories-1M.
In the layernorm layers we find mixed and very different results, but we value the layernorm
layers less because we mostly care about attention and MLP layers.

